\documentclass{article}
\PassOptionsToPackage{numbers, compress}{natbib}
\usepackage[preprint]{neurips_2026}
\usepackage[utf8]{inputenc} 
\usepackage[T1]{fontenc}    
\usepackage{hyperref}       
\usepackage{url}            
\usepackage{booktabs}       
\usepackage{amsfonts}       
\usepackage{nicefrac}       
\usepackage{microtype}      
\usepackage[table]{xcolor}  
\usepackage{multirow}
\usepackage{booktabs}
\usepackage{adjustbox}
\usepackage{dblfloatfix}
\usepackage{placeins}
\usepackage{caption}
\usepackage{amsmath}
\usepackage{amssymb}
\usepackage{subcaption}
\usepackage{tabularx}
\usepackage{array}
\usepackage{float}

\newcolumntype{L}[1]{>{\raggedright\arraybackslash}p{#1}}
\newcolumntype{Y}{>{\raggedright\arraybackslash}X}

\title{HADVQ: Hessian Aware Directional Vector Quantization}

\author{%
  \bfseries Daniil Li\textsuperscript{1,2},
  Mars Davletshin\textsuperscript{2},
  Vadim Klyzhenko\textsuperscript{2},
  Artem Danilishin\textsuperscript{2}\\
  \bfseries Valery Sandul\textsuperscript{2},
  Daniil Ryazanov\textsuperscript{2},
  Ekaterina Muravleva\textsuperscript{1}\\[0.5em]
  \normalfont \textsuperscript{1}Applied AI Institute, Moscow, Russia\\
  \normalfont \textsuperscript{2}Luzin Research Center, Moscow, Russia
}

\begin{document}

    \maketitle
    \begin{abstract}
    Ultra-low-bit quantization of large language models is a challenging task because aggressive compression severely distorts model parameters and disproportionately harms a small number of highly sensitive channels. We present a vector quantization method tailored for this regime. Our approach begins with random preconditioning, which regularizes the weight space and makes it more amenable to quantization. We then represent weights in polar coordinates and decouple the quantization of directions and magnitudes. To quantize directions, we leverage the structure induced by random preconditioning and design a structured codebook based on the $E_8$ lattice, yielding a compact yet expressive representation. To further reduce degradation, we detect extremely important channels using a proxy Hessian  and preserve them with an additional cost of only 0.16 bits. Together, these components provide an effective quantization pipeline for LLMs in the extremely low-bit regime. Empirical results demonstrate improved accuracy, highlighting the benefit of combining geometric structure, lattice-based codebooks, and selective channel preservation.
\end{abstract}
    \section{Introduction}

Large language models (LLMs) such as GPT~\citet{GPT}, LLaMA~\citet{llama, llama2, llama3}, DeepSeek, and Qwen~\citet{yang2025qwen3technicalreport} have become a central component of modern natural language processing. Their strong performance across a wide range of language tasks has led to rapid adoption, but their large memory footprint remains a major obstacle for practical deployment. For example, storing very large models in BF16 may require hundreds of gigabytes or even more than a terabyte of memory, which makes deployment expensive and hardware-demanding.

Several approaches have been proposed to reduce model size, including distillation~\citet{distil-1-intro, distil-2-intro}, pruning~\cite{pruning-1-intro}, low-rank approximation~\citet{low-rank-approximation-1-intro}, and quantization~\citet{quant-1-intro}. In this work, we focus on post-training quantization (PTQ), a practical approach for reducing the memory cost of pretrained models without retraining them from scratch. Scalar quantization (SQ)~\citet{frantar-gptq, lin2023awq, xiao2023smoothquant}, which maps each weight independently to a low-bit representation, is widely used and performs well in the 3--4-bit regime. However, in the more aggressive low-bit setting, vector quantization (VQ)~\citet{quip_original_2023, tseng2024quip, yue2025pcdvqenhancingvectorquantization} is often more robust because it can better capture local structure in the weight space.

In this paper, we study ultra-low-bit vector quantization \textbf{without fine-tuning}. Our method is motivated by the observation that quantization error is not distributed uniformly across channels: a relatively small subset of channels contributes disproportionately to performance degradation. Building on this observation, we propose a structured vector quantization method that combines random preconditioning, magnitude--direction decoupling, and Hessian-aware channel preservation.

\paragraph{Contributions.}
The main contributions of this work are as follows:
\begin{itemize}
    \item We propose a training-free vector quantization method that combines magnitude--direction decoupling with Hessian-aware preservation of sensitive channels.
    \item We introduce and study new strategies for directional codebook construction, and show in ablation experiments that the final design performs better than alternative initializations.
    \item We evaluate the method across multiple model families and benchmarks, including both question answering and perplexity tasks, and study how quantization sensitivity differs across models.
    \item We provide code and supporting materials to facilitate reproducibility.
\end{itemize}
    \section{Background}

\subsection{Post-training Quantization of LLMs}

Post-training quantization (PTQ) reduces the cost of a pretrained model by encoding its weights in a low-bit representation without retraining the model from scratch. For large language models, this setting is important because full-precision checkpoints use a lot of memory, and lower weight precision makes deployment easier.

PTQ for LLMs is difficult because quantization errors can accumulate across many layers and lead to a noticeable drop in model quality, as illustrated in Figure~\ref{fig:sensitive_channels_loss}. A common way to measure the local effect of quantization is to consider the expected output distortion induced by replacing a weight matrix $W$ with its quantized version $\hat W$, as in~\citet{pmlr-v119-nagel20a}
\begin{equation}
\ell(\hat W)
= \mathbb{E}_{x}\!\left[\|(\hat W - W)x\|^2\right]
= \operatorname{tr}\!\left((\hat W - W) H (\hat W - W)^T\right),
\label{eq:proxy_loss}
\end{equation}
where $H = \mathbb{E}[xx^T]$ is the input second-moment matrix.

This problem becomes more severe in the very low-bit regime. As a result, simple scalar quantization is often not sufficient at 2--3 bits~\cite{tseng2024quip, yue2025pcdvqenhancingvectorquantization}. In this work, we focus on weight-only PTQ with vector quantization in the extremely low-bit regime.

\subsection{Preconditioned Vector Quantization}

\paragraph{Vector Quantization.}
In our setting, weights are quantized in small blocks rather than independently, element by element. Let $W \in \mathbb{R}^{m \times n}$ be a weight matrix, and let each block be reshaped into a vector $w_i \in \mathbb{R}^d$. This allows us to use vector quantization (VQ), where each block is represented as a $d$-dimensional vector and replaced by the index of its nearest codeword from a codebook
\[
    C = \{c_1, \dots, c_K\}, \qquad c_j \in \mathbb{R}^d.
\]
The quantized block is then written as
\[
    Q(w_i)=\arg\min_{c\in C}\|w_i-c\|_2^2,
\]
If the quantization budget is $b$ bits per element, then each $d$-dimensional block is represented with $bd$ bits, and the codebook contains $K = 2^{bd}$ codewords.

Compared with scalar quantization, this gives the quantizer more flexibility to model correlations within each block, which is especially important in the extremely low-bit regime.

\paragraph{Random Preconditioning.}

A direct application of vector quantization to raw weight blocks is often difficult because their distribution is highly unbalanced across coordinates. In particular, a small number of coordinates may dominate the quantization error, which makes a fixed codebook less effective.

To make weight blocks more suitable for quantization, we first apply random preconditioning. Let $w \in \mathbb{R}^m$ be a weight column vector. We construct the preconditioned vector $\tilde w = H D w$, where $D \in \mathbb{R}^{m \times m}$ is a diagonal matrix with independent random signs on the diagonal, and $H \in \mathbb{R}^{m \times m}$ is the normalized Hadamard matrix. Since both $D$ and $H$ are orthogonal, this transformation preserves the Euclidean norm, that is, $\|\tilde w\|_2 = \|w\|_2$.

The role of this transformation is to spread the energy of the vector more evenly across coordinates. After random sign flipping and Hadamard mixing, each coordinate of $\tilde w$ becomes a sum of many signed components of $w$. As a result, the coordinates of $\tilde w$ become more balanced, and for sufficiently large block sizes, they can be viewed as approximately Gaussian due to a central-limit effect. More precisely, if we normalize by $\mathrm{scale} = \frac{\|w\|_2}{\sqrt{m}}$, then the coordinates of $\tilde w / \mathrm{scale}$ can be approximated by a standard Gaussian $\mathcal{N}(0,1)$. A proof is given in Appendix~\ref{appendix:rht_prove}.

After quantizing the preconditioned vector $\tilde w$, we obtain its low-bit approximation $\hat{\tilde w}$. This approximation can then be mapped back to the original space using the inverse transformation $w \approx D H^T \hat{\tilde w}$.

\paragraph{Magnitude–Direction Decoupling.}

After preconditioning, each weight block is represented as a vector $w \in \mathbb{R}^d$. We decompose $w$ into magnitude and direction
\[
w = r u, 
\qquad
r = \|w\|_2, 
\qquad
u = \frac{w}{\|w\|_2},
\]
where $r$ is the magnitude and $u$ is a unit vector with $\|u\|_2 = 1$.

This representation separates two different sources of quantization error. Errors in $r$ affect only the scale of the block, while errors in $u$ affect its direction. In the extremely low-bit regime, ~\citet{yue2025pcdvqenhancingvectorquantization} showed that directional distortion can be especially harmful. This makes it natural to quantize the two components independently.

The approximation $w \approx \hat w$ is written as $w = r u\longrightarrow\hat w = \hat r \hat u$,
where $\hat r$ and $\hat u$ are quantized separately. This allows the quantizer to allocate its limited bit budget more effectively: vector quantization can be focused on the directional component, while the magnitude can be handled with a simpler scalar quantizer.
    \section{HADVQ: Hessian Aware Directional Vector Quantization}

\begin{figure}
    \centering
    \includegraphics[width=\linewidth]{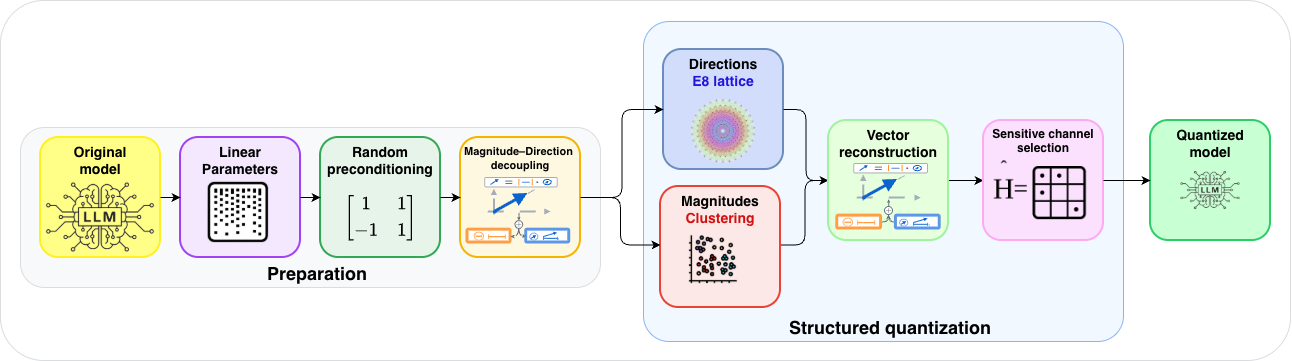}
    \caption{Pipeline of HADVQ. The method applies random preconditioning, decomposes each weight block into magnitude and direction, quantizes the two components separately, reconstructs the quantized blocks, and preserves highly sensitive channels using proxy Hessian scores.}
    \label{fig:hadvq_pipeline}
\end{figure}

HADVQ combines several ideas within a single post-training quantization pipeline. Figure~\ref{fig:hadvq_pipeline} gives an overview of the method. Starting from the weights of linear layers in a pretrained model, we first apply random preconditioning to make weight blocks more suitable for vector quantization. Each preconditioned block is then represented by its magnitude and direction. The directional component is quantized with a structured codebook, while the magnitude is quantized separately. The quantized direction and magnitude are then combined to reconstruct the weight block. Finally, to reduce the degradation caused by aggressive compression, we preserve a small subset of highly sensitive channels selected using proxy Hessian scores. Together, these steps define our method for post-training quantization in the extremely low-bit regime.

\subsection{Codebooks construction}

\paragraph{Directional codebook construction.}
To build a structured directional codebook in $\mathbb{R}^8$, we use normalized points from shells of the $E_8$ lattice. For a shell of radius $R$, we define
\[
    \Lambda_R = \{x \in E_8 : \|x\|_2 = R\},
    \qquad
    \mathcal{C}_R = \left\{ \frac{x}{R} : x \in \Lambda_R \right\} \subset S^7 ,
\]
where $S^7 = \{u \in \mathbb{R}^8 \mid \|u\|_2 = 1\}$ is the unit sphere in $\mathbb{R}^8$. Thus, each codeword is obtained by normalizing a lattice point on the shell of radius $R$, and the resulting codebook consists of unit directions.

This construction is well matched to our preconditioned setting. After random Hadamard preconditioning, the coordinates of weight blocks become approximately Gaussian, as also observed in PolarQuant~\cite{1051813}. As a consequence, the normalized directions can be viewed as approximately uniform on the sphere. This makes normalized shells of the $E_8$ lattice a natural choice for the directional codebook, since they have strong uniformity properties. In particular,~\citet{bannai2009survey} and~\citet{elkies2019e8shells} show that each normalized shell of the $E_8$ lattice forms a spherical $7$-design. This means that the resulting directions are distributed in a highly symmetric and well-balanced way, which makes them suitable as a structured directional codebook.

The number of $E_8$ lattice points with squared norm $n$ is
\begin{equation}
N_{E_8}(n)=
\begin{cases}
1, & n = 0,\\[4pt]
240\,\sigma_3(n/2), & n \in 2\mathbb{Z}_{>0},\\[4pt]
0, & \text{otherwise},
\end{cases}\label{eq:num_points}
\end{equation}
where $\sigma_3(n)=\sum_{d\mid n} d^3$. Therefore, a shell of radius $R$ is nonempty only if $R^2$ is a positive even integer.

In practice, the shell size does not always match the target codebook size of $2^{b\cdot d}$. To obtain the final codebook, we augment the lattice-based codebook with additional directions taken from weight blocks with the largest directional approximation error. For a block with magnitude $r$, original direction $u$, and quantized direction $\hat u$, this squared directional reconstruction error is
\[
\|ru-r\hat u\|_2^2 = 2r^2(1-\cos\theta),
\]
where $\theta$ is the angle between $u$ and $\hat u$. We rank candidate directions by this score and add those with the largest values until the desired codebook size is reached.

\paragraph{Magnitude codebook construction.}
After magnitude--direction decoupling, each weight block is written as $w = ru$, where $r=\|w\|_2$ is the magnitude. Since $r$ is a scalar, we quantize it separately using a one-dimensional codebook. Following PCDVQ~\cite{yue2025pcdvqenhancingvectorquantization}, we construct this codebook with the Lloyd--Max algorithm~\cite{lloyd}, which iteratively minimizes the mean squared reconstruction error over the empirical distribution of magnitudes. Given a codebook $C_m=\{c^{1}_{m},\dots,c_{m}^{K}\}$ with $K=2^{b_m}$ quantization levels, each magnitude is quantized as
\[
    Q_r(r)=\arg\min_{c\in C_m}|r-c|^2.
\]

\subsection{Hessian-Aware Channel Preservation}

A small subset of input channels can contribute disproportionately to the total quantization error. To reduce this effect, we preserve the most sensitive input channels in higher precision and quantize only the remaining channels.

We estimate channel sensitivity based on the proxy loss in Eq.~\eqref{eq:proxy_loss}. In our setting, a channel corresponds to an input dimension of a linear weight matrix $W \in \mathbb{R}^{m \times n}$. Let $h \in \mathbb{R}^n$ denote the diagonal of the proxy Hessian, and let $w_j \in \mathbb{R}^m$ be the $j$-th column of $W$, corresponding to the $j$-th input channel. We define the sensitivity score of this channel as
\[
s_j = h_j \|w_j\|_2^2.
\]
In vectorized form, this is computed as
\[
s = h \odot \Bigl(\sum\nolimits_{i=1}^{m} W_{i,:}^2\Bigr),
\]
where $\odot$ denotes element-wise multiplication. We rank input channels according to these scores and preserve the top-ranked fraction in higher precision. Although the preserved fraction is small, this step noticeably improves model quality in the extremely low-bit regime.

    \section{Experimental Results}

\subsection{Settings}

\paragraph{Quantization.} The directional component is quantized using an $E_8$-based vector codebook. Specifically, we use shells with radii $\sqrt{12}$, $\sqrt{14}$, and $\sqrt{78}$, which yield 60\,480, 82\,560, and 14\,770\,560 candidate directions, respectively, as given by Eq.~\ref{eq:num_points}. These codebooks are then expanded to sizes $2^{16}$, $2^{17}$, and $2^{24}$ to obtain integer directional bit-widths. The magnitude is quantized independently to 2 bits using the Lloyd--Max algorithm~\cite{lloyd}. We also preserve the top $1\%$ of highly sensitive channels.

\paragraph{Bit budget.} 
If the directional component uses $b_d$ bits per block, the magnitude uses $b_m$ bits per block, the block size is $d$, and a fraction $p$ of highly sensitive channels is preserved in precision $t$, then the average number of bits per parameter is
\[
b_{\text{avg}} = \frac{b_d + b_m}{d} + t\,p.
\]
In our experiments, we use $d=8$, $b_m=2$, $t=16$, and $p=0.01$. This makes $b_{\text{avg}}$ equal to $2.41$, $2.535$, and $3.41$, respectively, for different codebook sizes.

\paragraph{Models and Datasets.} We evaluate a diverse collection of large language models from the Qwen3~\cite{yang2025qwen3technicalreport}, Llama2~\cite{llama2}, and Llama3~\cite{llama3} families. Our experiments cover sizes ranging from 0.6B to 13B parameters, including Qwen3-0.6B/1.7B/4B/8B, Llama2-7B/13B, Llama3-8B, and Llama3.2-1B/3B. We report perplexity on the WikiText-2~\cite{merity2016pointer} test set and zero-shot accuracy on BoolQ~\cite{clark2019boolq}, WinoGrande~\cite{sakaguchi2019winogrande}, ARC-Easy and ARC-Challenge~\cite{Clark2018ThinkYH}, HellaSwag~\cite{zellers2019hellaswag}, PIQA~\cite{Bisk2020}, and MMLU~\cite{hendryckstest2021, hendrycks2021ethics}, and additionally report five-shot results on MMLU. For downstream evaluation, we use the LM Eval Harness~\cite{eval-harness} versions 0.3.0, 0.4.4, and 0.4.11, depending on model and benchmark compatibility.

\paragraph{Baselines.} We compare against representative PTQ baselines. Scalar quantization baselines include GPTQ~\cite{frantar-gptq}, AWQ~\cite{lin2023awq}, QuaRot~\cite{ashkboos2024quarot}, OSTQuant~\cite{hu2025ostquant}, and ReSpinQuant~\cite{kim2026respinquantefficientlayerwisellm}. Vector quantization baselines include QuIP\#~\cite{tseng2024quip}, AQLM~\cite{egiazarian2024extreme}, ICQuant~\cite{li2025icquant}, VPTQ~\cite{vptq}, PCDVQ~\cite{yue2025pcdvqenhancingvectorquantization}, and QTIP~\cite{tseng2024qtip}. Some vector quantization methods, such as QuIP\#, AQLM, PCDVQ, and QTIP, can optionally be combined with fine-tuning to further improve performance. In contrast, our method is evaluated in a pure post-training setting without fine-tuning. We leave this direction for future work.

\paragraph{Calibration setup.} For proxy Hessian estimation, we use \texttt{mit-han-lab/pile-val-backup}~\cite{gao2020pile800gbdatasetdiverse} with 256 calibration samples of sequence length 1024. The resulting scores are used to identify highly sensitive channels for selective preservation.

\paragraph{Implementation details.} We use a block size of 8 and evaluate effective bit budgets of 2.41, 2.535, and 3.41 bits per parameter. We quantize directional and magnitude components separately. Magnitudes are quantized to 2 bits using the Lloyd--Max algorithm~\cite{lloyd}, while directions are quantized to 2, 2.125, and 3 bits for the 2.41, 2.535, and 3.41 bit settings, respectively. For efficient nearest-neighbor lookup in the directional codebook, we use \texttt{torch\_kdtree}~\cite{grandits_geasi_2021}. Randomized Hadamard preconditioning is implemented with \texttt{fast-hadamard-transform}~\cite{fast_hadamard_transform}. Proxy Hessian scores are aggregated at the channel level and used to rank channels by sensitivity, and the top-ranked channels are preserved separately. Our implementation is based on PyTorch~\cite{paszke2019pytorchimperativestylehighperformance} and Hugging Face Transformers~\cite{wolf-etal-2020-transformers}. All experiments were conducted on a single V100 32GB GPU.

\subsection{Main Results}

Table~\ref{tab:llama_main} reports downstream results for Llama2-7B, Llama2-13B, and Llama3-8B. Overall, HADVQ provides a strong accuracy-compression trade-off across all three models and remains competitive with both scalar and vector quantization baselines in the $\sim$2-bit and $\sim$3-bit regimes.

At around 3 bits, HADVQ preserves model quality well and achieves the best results on several tasks across all three models. In the more aggressive low-bit setting, performance drops as expected, but HADVQ remains stable and competitive with strong baselines such as PCDVQ, QTIP, and ICQuant. This is especially notable since several competing methods, marked by square brackets, use fine-tuning, whereas HADVQ is evaluated purely in the post-training setting.

Table~\ref{tab:wiki2_llama_main} shows a similar pattern for WikiText-2 perplexity at sequence lengths 2048 and 4096. In the $\sim$3-bit regime, HADVQ consistently matches or improves upon prior methods across all three models, while in the lower-bit regime, it remains substantially more stable than several aggressive baselines. Additional results are provided in the Appendix~\ref{appendix:additional_results}.

\begin{table}[h!]
    \centering
    \caption{Zero-shot accuracy ($\uparrow$) comparison on Llama2-7B, Llama2-13B, and Llama3-8B under low-bit quantization. Gray rows denote our method. $[\cdot]$ indicates results obtained with any form of fine-tuning. All results are reported using LM Eval Harness v0.3.0.}
    \small
    \scriptsize
    \setlength{\tabcolsep}{2.5pt}
    \renewcommand{\arraystretch}{0.92}
        \begin{tabular}{llcccccccc}
            \toprule
            \textbf{Model} & \textbf{Method} & \textbf{Bits} & \textbf{BoolQ}$\uparrow$ & \textbf{WinoGrande}$\uparrow$ & \textbf{ArcE}$\uparrow$ & \textbf{ArcC}$\uparrow$ & \textbf{HellaSwag}$\uparrow$ & \textbf{PiQA}$\uparrow$ \\
            \midrule
            \multirow{17}{*}{Llama2-7B}
            & FP16    & 16   & 71.10 & 67.17 & 69.28 & 39.93 & 56.70 & 78.35 \\
            & QuIP\#  & 3    & [69.70] & [66.50] & [68.40] & \textbf{[39.20]} & - & [77.30] \\
            & AQLM    & 3.04 & - & \textbf{[66.93]} & [68.06] & [38.40] & [54.12] & [76.88] \\
            & QTIP    & 3    & - & [66.90] & [68.10] & [38.90] & - & \textbf{[78.10]} \\
            & ICQuant & 3.3  & - & 66.90 & 66.60 & 39.00 & - & 77.70 \\
            & \cellcolor{gray!15} HADVQ
            & \cellcolor{gray!15} 3.41
            & \cellcolor{gray!15}\textbf{70.95}
            & \cellcolor{gray!15}66.85
            & \cellcolor{gray!15}\textbf{69.07}
            & \cellcolor{gray!15}38.48
            & \cellcolor{gray!15}\textbf{55.49}
            & \cellcolor{gray!15}77.97 \\

            & QuIP\#  & 2     & [68.30] & [64.90] & [64.60] & [34.60] & - & [75.10] \\
            & AQLM    & 2.02  & - & [65.67] & [63.68] & [32.76] & [49.55] & [74.76] \\
            & AQLM    & 2.29  & - &	[65.67] & [66.50] & [34.90] & [50.88] &	[74.92] \\
            & QTIP    & 2     & - & [64.70] & [65.60] & [35.70] & - & [75.90] \\
            & ICQuant & 2.3   & - &	63.60 &	62.00 & 34.10 & - &	75.00 \\
            & ICQuant & 2.4   & - & 65.60 & 63.60 & 35.90 & - & 76.10 \\
            & VPTQ  & 2.02  & - & [64.33] & [63.80] & [35.24] & [52.08] & [75.19] \\
            & VPTQ  & 2.26  & - & [66.46] & [64.90] & [36.43] & [52.87] & [76.17] \\
            & PCDVQ   & 2.25 & - & [66.46] & \textbf{[67.92]} & \textbf{[38.31]} & \textbf{[53.13]} & [76.38] \\
            & \cellcolor{gray!15} HADVQ
            & \cellcolor{gray!15} 2.535
            & \cellcolor{gray!15}\textbf{72.69}
            & \cellcolor{gray!15}65.35
            & \cellcolor{gray!15}66.96
            & \cellcolor{gray!15}35.49
            & \cellcolor{gray!15}52.86
            & \cellcolor{gray!15}75.95 \\
            & \cellcolor{gray!15} HADVQ
            & \cellcolor{gray!15} 2.41
            & \cellcolor{gray!15}69.02
            & \cellcolor{gray!15}\textbf{66.54}
            & \cellcolor{gray!15}65.74
            & \cellcolor{gray!15}36.35
            & \cellcolor{gray!15}51.55
            & \cellcolor{gray!15}\textbf{76.88} \\

            \midrule
            \multirow{16}{*}{Llama2-13B}
            & FP16      & 16      & 68.96     & 69.69   & 73.23   & 45.56   & 59.74   & 78.78       \\
            & QuIP\#    & 3       & \textbf{[69.80]}	  & [69.10] & [69.30] & [39.50] & -       & [77.30]     \\
            & AQLM      & 3.03    & -         & [68.43] & [70.88] & [42.58] & [58.30] & [77.26]     \\
            & QTIP      & 3       & -         & [69.50] & \textbf{[72.80]} & [44.00] & -       & [72.80]     \\
            & ICQuant   & 3.3     & -       	 & 69.80   & 70.90   & \textbf{44.50}   & -       & \textbf{78.70}    \\
            & \cellcolor{gray!15} HADVQ
            & \cellcolor{gray!15} 3.41
            & \cellcolor{gray!15}69.05
            & \cellcolor{gray!15}\textbf{70.24}
            & \cellcolor{gray!15}72.69
            & \cellcolor{gray!15}43.43
            & \cellcolor{gray!15}\textbf{58.86}
            & \cellcolor{gray!15}78.45 \\

            & QuIP\#        &   2       &   [68.30] &	[67.70]   &	\textbf{[72.50]}   &	\textbf{[44.00]}   &   -       &	\textbf{[78.40]}        \\
            & AQLM          &   1.97    &   -       &	[68.82]   &	[69.28]  &	[38.82]  &	[53.80]  &	[75.90]       \\
            & AQLM          &   2.19    &   -       &	[67.64]  & [70.29]    & [38.65]    &   [55.03]    &	[77.37]  \\
            & AQLM          &   2.53    &   -       & 	[69.77] &   [70.33]  &	[39.16]&	[56.15]&	[76.99] \\
            & QTIP      & 2 & - & 	[67.60]&	[70.80]&	[41.40]& - &		[70.80] \\
            & ICQuant & 2.3 & - &	64.70 &	69.10 &	39.40 & - &	76.90 \\
            & ICQuant & 2.4 & - & 	64.20 &	68.40 &	39.80 &	- &	77.00 \\
            & VPTQ & 2.02 & - & [66.85] & [71.55] & [40.02] & [56.18] & [77.26] \\
            & VPTQ & 2.18 & - & [68.43] & [71.80] & [40.96] & [56.89] & [77.28] \\
            & PCDVQ & 2.25 & - &	\textbf{[70.63]} &	[70.53] &	[43.34] & 	[56.57] &	[77.25] \\
            & \cellcolor{gray!15} HADVQ
            & \cellcolor{gray!15} 2.535
            & \cellcolor{gray!15}\textbf{72.05}
            & \cellcolor{gray!15}66.77
            & \cellcolor{gray!15}69.82
            & \cellcolor{gray!15}40.61
            & \cellcolor{gray!15}\textbf{56.71}
            & \cellcolor{gray!15}77.64\\
            & \cellcolor{gray!15} HADVQ
            & \cellcolor{gray!15} 2.41
            & \cellcolor{gray!15}75.20
            & \cellcolor{gray!15}67.80
            & \cellcolor{gray!15}69.87
            & \cellcolor{gray!15}36.18
            & \cellcolor{gray!15}49.43
            & \cellcolor{gray!15}74.65\\

            \midrule
            \multirow{14}{*}{Llama3-8B}
            & FP16    & 16   & 81.38 & 72.69 & 80.09 & 50.43 & 60.17 & 79.65 \\
            & Quip\#  & 3    & - & [72.90] & [77.40] & [46.40] & - & [77.90] \\
            & QTIP    & 3    & - & [74.50] & [79.30] & [49.20] & - & [79.20] \\
            & ICQuant & 3.3  & - & 72.90 & 78.10 & 47.40 & - & 78.60 \\
            & \cellcolor{gray!15} HADVQ
            & \cellcolor{gray!15} 3.41
            & \cellcolor{gray!15}\textbf{81.38}
            & \cellcolor{gray!15}71.74
            & \cellcolor{gray!15}\textbf{77.99}
            & \cellcolor{gray!15}\textbf{47.01}
            & \cellcolor{gray!15}\textbf{56.21}
            & \cellcolor{gray!15}\textbf{78.29} \\
            & GPTQ    & 2     & - & 50.50 & 28.80 & 19.90 & 27.70 & 53.90 \\
            & VPTQ    & 2.08  & - & [65.90] & [71.00] & [36.90] & [52.20] & [75.10] \\
            & VPTQ    & 2.24  & - & [69.10] & [73.20] & [42.60] & [53.10] & [75.40] \\
            & PCDVQ   & 2.25  & - & \textbf{[70.08]} & \textbf{[75.04]} & \textbf{[43.68]} & \textbf{[54.29]} & \textbf{[76.49]} \\
            & Quip\#  & 2     & - & [68.20] & [72.90] & [39.20] & - & [75.60] \\
            & QTIP    & 2     & - & [70.70] & [75.20] & [44.20] & - & [77.60] \\
            & ICQuant & 2.4   & - & 67.90 & 71.60 & 38.00 & - & 76.60 \\
            & ICQuant & 2.3   & - & 66.60 & 70.70 & 35.80 & - & 74.90 \\
            & \cellcolor{gray!15} HADVQ
            & \cellcolor{gray!15} 2.535
            & \cellcolor{gray!15}\textbf{75.96}
            & \cellcolor{gray!15}68.75
            & \cellcolor{gray!15}72.64
            & \cellcolor{gray!15}41.81
            & \cellcolor{gray!15}51.50
            & \cellcolor{gray!15}75.90 \\
            & \cellcolor{gray!15} HADVQ
            & \cellcolor{gray!15} 2.41
            & \cellcolor{gray!15}\textbf{75.20}
            & \cellcolor{gray!15}67.80
            & \cellcolor{gray!15}69.87
            & \cellcolor{gray!15}36.18
            & \cellcolor{gray!15}49.43
            & \cellcolor{gray!15}74.65 \\
            \bottomrule
        \end{tabular}
    \label{tab:llama_main}
\end{table}[h!]

\begin{table}[h!]
\centering
\caption{WikiText-2 perplexity ($\downarrow$) for sequence lengths 2048 and 4096. $[\cdot]$ indicates results reported with fine-tuning.}
\label{tab:wiki2_llama_main}
\scriptsize
\setlength{\tabcolsep}{3pt}
\renewcommand{\arraystretch}{0.9}
\begin{tabular}{l ccc ccc ccc}
\toprule
& \multicolumn{3}{c}{Llama2-7B} & \multicolumn{3}{c}{Llama2-13B} & \multicolumn{3}{c}{Llama3-8B} \\
\cmidrule(lr){2-4} \cmidrule(lr){5-7} \cmidrule(lr){8-10}
\textbf{Method} & \textbf{bits} & \textbf{2048}$\downarrow$ & \textbf{4096}$\downarrow$ & \textbf{bits} & \textbf{2048}$\downarrow$ & \textbf{4096}$\downarrow$ & \textbf{bits} & \textbf{2048}$\downarrow$ & \textbf{4096}$\downarrow$ \\
\midrule
FP16        & 16    & 5.47   & 5.12   & 16  & 4.88   & 4.57   & 16  & 6.14   & 5.75 \\
QuIP\#      & 3     & [5.79] & [5.41] & 3   & [5.10] & [4.78] & --  & --     & --   \\
AQLM        & 3.04  & --     & [5.46] & 3.03    & --     & [4.82]     & --  & --     & --   \\
QTIP        & 3     & --     & \textbf{[5.28]} & 3   & --     & \textbf{[4.69]}     & --   & --    & --  \\
ICQuant     & 3.3   & --     & 5.35   & 3.3 & --     & 4.75   & --  & --     & --   \\
VPTQ        & 3.02  & [5.82]    & [5.43]    & 3.03  & [5.12]  & [4.79]  & 3.03 & [6.97] & -- \\
\rowcolor{gray!15}
HADVQ       & 3.41  & \textbf{5.72} & 5.35 & 3.41    & \textbf{5.07} & 4.74 & 3.41  & \textbf{6.92} & \textbf{6.48} \\

ICQuant     & 2.3   & 7.21   & 6.75   & 2.3 & 6.10   & 5.74   & -- & --     & --   \\
ICQuant     & 2.4   & --     & 6.35   & 2.4 & --     & 5.54   & -- & --     & --   \\

VPTQ        & 2.02  & --    & [6.13]    & 2.02  & --    & [5.32] & 2.08 & 9.29  & -- \\
VPTQ        & 2.26  & --    & [5.95]    & 2.18  & --    & [5.28] & 2.24 & 9.19  & -- \\

AQLM        & 2.02  & --    & [6.59]    & 1.97  & --    & [5.60] & --   & -- & -- \\
AQLM        & 2.29  & --    & [6.29]    & 2.76  & --    & \textbf{[4.94]} & --   & -- & -- \\

QuIP\#      & 2     & [6.66] & [6.19] & 2   & [5.74] & [5.35] & --  & --     & --   \\

QTIP        & 2     & --     & [5.86] & 2   & --     & [5.11]     & --  & --     & --   \\
PCDVQ       & 2.225 & --     & \textbf{[5.68]} & 2.225  &  --     & [5.04] & 2.225 & \textbf{[7.93]} & --   \\
\rowcolor{gray!15}
HADVQ       & 2.535 & \textbf{6.42}   & 5.98   & 2.535  & \textbf{5.52}   & 5.16   & 2.535  & 9.03   & \textbf{8.46} \\
\rowcolor{gray!15}
HADVQ       & 2.41  & 6.65   & 6.20   & 2.41    & 5.69   & 5.31   & 2.41    & 9.86   & 9.21 \\

\bottomrule
\end{tabular}
\end{table}

\subsection{Ablation Study}

\begin{figure}[t]
    \centering
    \begin{subfigure}{0.49\linewidth}
        \centering
        \includegraphics[width=\linewidth]{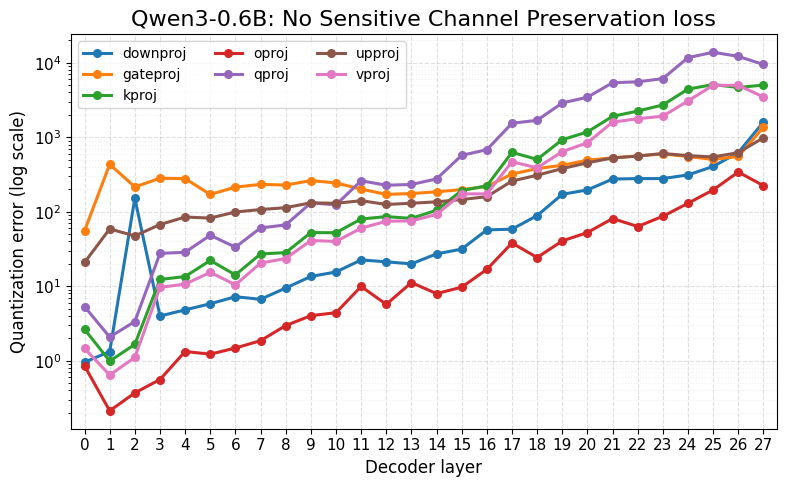}
        \caption{Without sensitive-channel preservation.}
    \end{subfigure}
    \hfill
    \begin{subfigure}{0.49\linewidth}
        \centering
        \includegraphics[width=\linewidth]{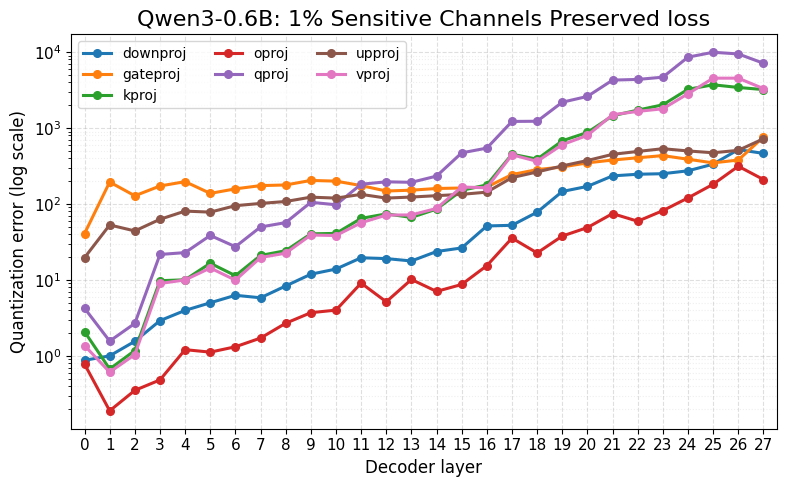}
        \caption{With 1\% sensitive channels preserved.}
    \end{subfigure}
    
    \caption{Layer-wise quantization loss on Qwen3-0.6B for different projection matrices.}
    \label{fig:sensitive_channels_loss}
\end{figure}

We study two aspects of the proposed method. First, we evaluate the contribution of selective preservation of highly sensitive channels and the proposed directional codebook construction. Second, we examine how sensitive the set of selected highly sensitive channels is to the choice of calibration dataset.

\begin{figure}[h!]
\centering

\begin{minipage}[t]{0.54\linewidth}
    \vspace{0pt}
    \centering
    \includegraphics[width=0.82\linewidth]{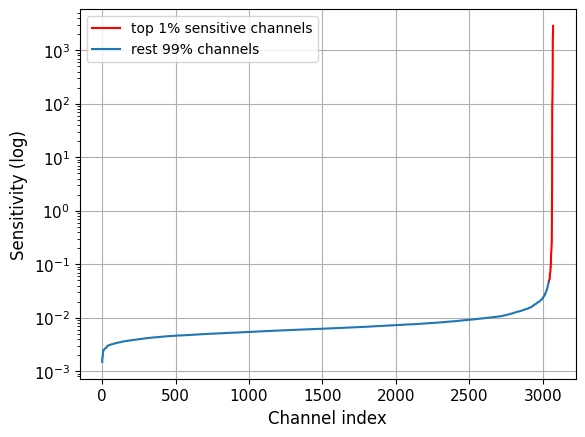}
    \caption{Sensitive channels for model Qwen3-0.6B in \texttt{layers.2.mlp.down\_proj} layer.}
    \label{fig:sensitive_channels}
\end{minipage}
\hfill
\begin{minipage}[t]{0.40\linewidth}
    \vspace{0pt}
    \centering
    \captionsetup{type=table}
    \caption{Overlap ratio of the selected top 1\% highly sensitive channels for Qwen3-1.7B across different calibration datasets.}
    \label{tab:channel_overlap}
    \small
    \setlength{\tabcolsep}{8pt}
    \renewcommand{\arraystretch}{1.0}
    \begin{tabular}{lc}
    \toprule
    \textbf{Dataset pair} & \textbf{Overlap ratio (\%)} \\
    \midrule
    The Pile / C4    & 91.7 \\
    The Pile / BoolQ & 78.7 \\
    BoolQ / C4       & 80.4 \\
    \bottomrule
    \end{tabular}
\end{minipage}

\end{figure}

\paragraph{Impact of sensitive channels and codebook design.}
Table~\ref{tab:ablation_main} reports ablations on Qwen3-0.6B, Qwen3-4B, and Qwen3-8B. We compare the full HADVQ pipeline against several variants: removing sensitive-channel preservation, replacing the selected channels with a random subset of the same size, and replacing the proposed directional codebook construction with alternative initialization strategies. We observe the same pattern across all three models. First, preserving a small subset of highly sensitive channels is important in the extremely low-bit regime: removing these channels or replacing them with random ones leads to a clear drop in both downstream accuracy and WikiText-2 perplexity. Figure~\ref{fig:sensitive_channels_loss} shows that preserving 1\% of sensitive channels reduces quantization loss on a number of layers, although it does not remove the overall growth of the loss across depth and does not fully stop error accumulation. Figure~\ref{fig:sensitive_channels} shows that 99\% of channels have relatively small sensitivity compared to the top 1\% sensitive channels. Second, the proposed codebook construction is also important: both the greedy min-max initialization, similar to the procedure used in~\citet{yue2025pcdvqenhancingvectorquantization}, and the evolutionary alternative perform worse than the final HADVQ design. For the evolutionary approach, we use the method of~\citet{khrulkov2025gigaevoopensourceoptimization}. The best results are obtained when both components are used together. We also observe that reducing the preserved fraction from 1\% to 0.5\% only slightly affects performance on Qwen3-4B and Qwen3-8B, but causes a more visible drop on Qwen3-0.6B, suggesting that smaller models are more sensitive to this budget. 

\begin{table}[h!]
\centering
\caption{Ablation of sensitive-channel preservation and directional codebook construction. We report average zero-shot accuracy (\%) over BoolQ, WinoGrande, ARC-Easy, ARC-Challenge, and PIQA, together with WikiText-2 perplexity. Results were obtained with LM Eval Harness v0.4.11.}
\label{tab:ablation_main}
\scriptsize
\setlength{\tabcolsep}{4pt}
\renewcommand{\arraystretch}{1.02}
\begin{tabular}{l c cc cc cc}
\toprule
& & \multicolumn{2}{c}{Qwen3-0.6B} & \multicolumn{2}{c}{Qwen3-4B} & \multicolumn{2}{c}{Qwen3-8B} \\
\cmidrule(lr){3-4} \cmidrule(lr){5-6} \cmidrule(lr){7-8}
\textbf{Variant} & \textbf{Bits} & \textbf{Avg. Acc}$\uparrow$ & \textbf{Wiki2}$\downarrow$ & \textbf{Avg. Acc}$\uparrow$ & \textbf{Wiki2}$\downarrow$ & \textbf{Avg. Acc}$\uparrow$ & \textbf{Wiki2}$\downarrow$ \\
\midrule
FP16 & 16 & 56.05 & 20.90 & 71.40 & 13.70 & 73.98 & 9.71 \\
\rowcolor{gray!15}
HADVQ (1\% important, final codebook) & 2.41 & 45.73 & 49.39 & 64.09 & 17.73 & 69.95 & 12.10 \\
w/o important channels & 2.25 & 45.47 & 107.71 & 61.25 & 21.34 & 64.60 & 14.18 \\
1\% random channels & 2.41 & 43.90 & 249.04 & 59.19 & 24.56 & 48.04 & 69.98 \\
greedy min--max codebook & 2.25 & 40.45 & 306.02 & 58.40 & 26.69 & 51.82 & 55.53 \\
evolutionary codebook & 2.25 & 39.67 & 29013.72 & 52.51 & 36.14 & 49.18 & 53.17 \\
\rowcolor{gray!15}
HADVQ (0.5\% important, final codebook) & 2.33 & 46.11 & 56.74 & 63.14 & 17.84 & 69.58 & 12.41 \\
\bottomrule
\end{tabular}
\end{table}

\paragraph{Sensitivity to the calibration dataset.}
We further study whether the selection of highly sensitive channels depends strongly on the calibration dataset. For this experiment, we use Qwen3-1.7B and compute the top 1\% most important channels using three different calibration datasets: Pile~\cite{gao2020pile800gbdatasetdiverse}, BoolQ~\cite{clark2019boolq}, and C4~\cite{raffel2023exploringlimitstransferlearning}. Table~\ref{tab:channel_overlap} reports the overlap ratio between the resulting channel sets.

The overlap remains high across all dataset pairs: 91.7\% for Pile vs. C4, 78.7\% for  Pile vs. BoolQ, and 80.4\% for BoolQ vs. C4. This suggests that the ranking of highly sensitive channels is fairly stable with respect to the calibration data and that the selected channels are not determined by a single narrow dataset. Based on this observation, we use The Pile as the default calibration dataset.


    \section{Related Work}

\paragraph{Post-Training Quantization.} PTQ is an approach for compressing LLMs without retraining, and it has become a crucial technique for LLMs deployment. \citet{lin2023awq} introduced a technique that protects a small subset of salient channels determined from the activation distribution. \citet{frantar-gptq} showed a method that is based on approximate second-order information. \citet{xiao2023smoothquant} explored the fact that weights are easy to quantize while activations are not; they smoothed the activation outliers from activations to weights. These works show that the quality of the PTQ method depends on salient channels, second-order information, and outlier handling.

\paragraph{Sensitivity-aware and outlier-aware quantization.} An important observation in quantization is that the sensitivity of weight channels and outliers of weight values can dramatically influence quantization performance. There are works~\cite{dong2019hawqhessianawarequantization,dong2019hawqv2hessianawaretraceweighted,yao2021hawqv3dyadicneuralnetwork} that are dedicated to measuring layer sensitivity using Hessian trace information. As mentioned above~\citet{lin2023awq} used activation statistics to identify salient channels.~\citet{li2025icquant} leveraged outlier statistics to design an efficient index coding scheme for low-bit quantization. Our method belongs to this family of methods, we identify an extremely small subset of highly sensitive channels using a proxy Hessian criterion.

\paragraph{Per-channel transformations.} A single global quantization scale does not consider sensitive channels and different value ranges. Therefore, per-channel transformations improve quantization quality. \citet{xiao2023smoothquant} uses per-channel scaling to migrate outlier channels from activation to weight. \citet{OmniQuant} uses a differentiable approach to learn optimal
scaling and shifting parameters. In our work, we use another approach with random preconditioning to change the geometry of the weight space.

\paragraph{Preconditioning-based quantization.} Another line of work improves quantization by transforming the geometry of weights and activations before the quantization procedure. \citet{sun2024flatquant} developed a learnable affine transformation that flattens the distributions of weights and activations. \citet{kim2026respinquantefficientlayerwisellm} proposed a rotation-based transformation using offline activation rotation fusion and matching bases with efficient residual subspace rotation. \citet{tseng2024quip} uses randomized Hadamard preconditioning to make weight matrices more incoherent. These methods explain that the geometry of weights/activations is a very important factor for low-bit quantization.

\paragraph{Vector quantization.} In ultra-low-bit scenario, scalar quantization becomes inadequate because it is difficult to consider the distribution and geometry of weights. \citet{egiazarian2024extreme} demonstrated from the point of view of additive quantization and multi-codebook quantization through learned additive codebooks. \citet{tseng2024quip} combines incoherence processing through random preconditioning with 
lattice codebooks based on the $E_8$ lattice. \citet{vptq} formulates vector post-training quantization as a second-order optimization problem and introduces channel-independent refinement and improved codebook initialization. \citet{yue2025pcdvqenhancingvectorquantization} found out that the directional component of a weight vector exhibits significantly greater quantization sensitivity compared to magnitude. Our method is  closely related to this line of work; it combines random preconditioning, direction-magnitude decomposition, and structured vector quantization.

    \section{Conclusion}

We presented HADVQ, a post-training quantization method for large language models in the extremely low-bit regime. The method combines random preconditioning, magnitude--direction decoupling, structured directional vector quantization, and Hessian aware preservation of sensitive input channels. Together, these components provide a practical way to reduce model size while maintaining strong downstream performance.

Across multiple Llama and Qwen models, HADVQ achieves a favorable accuracy--compression trade-off and remains competitive with recent scalar and vector quantization baselines, despite not using fine-tuning. The ablation study further shows that both the structured directional codebook and sensitive-channel preservation contribute to the final performance, while the selected channels remain relatively stable across calibration datasets.

A direction for future work is to extend HADVQ beyond the current post-training setting, for example by combining it with fine-tuning or by studying larger block sizes and alternative structured codebooks.
    \section{Limitations}

This work has several limitations. First, we consider only the weight-only post-training quantization setting and do not study activation quantization or joint weight--activation compression. Second, the current method is evaluated purely without fine-tuning, so its behavior in combination with post-quantization fine-tuning remains an open question. Third, the directional codebook is based on the $E_8$ lattice and is therefore naturally tied to 8-dimensional blocks; extending the same construction to other block sizes may require different structured codebooks. Finally, the method relies on a random Hadamard transform (RHT), which must also be applied during de-quantization and therefore introduces additional runtime overhead. In practice, this overhead is moderate since the transform can be computed in $O(n\log n)$ time. The additional memory used by the codebooks and the sensitive-channel mask is shared across the whole model and is therefore negligible when spread over billions of parameters.
    \bibliographystyle{plainnat}
    \bibliography{bibliography}

    \newpage
    \appendix
    \section{Proof of Approximate Gaussianity under Randomized Hadamard Preconditioning}\label{appendix:rht_prove}

\paragraph{Lemma.}
Let $w \in \mathbb{R}^m$ be a deterministic vector such that
\[
\frac{\|w\|_\infty}{\|w\|_2} \xrightarrow[m\to\infty]{} 0.
\]
Let $D = \mathrm{diag}(d_1, \dots, d_m)$ with $d_j \overset{i.i.d.}{\sim} \mathrm{Unif}\{-1,1\}$, and let $H \in \mathbb{R}^{m \times m}$ be a normalized Hadamard matrix, $H H^\top = I$, $h_{ij}^2 = 1/m$. Define
\[
\tilde w = H D w.
\]
Then, for any fixed coordinate $i$,
\[
\frac{\sqrt{m}\,\tilde w_i}{\|w\|_2}
\;\xrightarrow{d}\;
\mathcal{N}(0,1).
\]

\paragraph{Proof.}
Write $\tilde w_i = \sum_{j=1}^m \xi_j^{(i)}$, where $\xi_j^{(i)} := h_{ij} d_j w_j$ are independent, $\mathbb{E}\xi_j^{(i)} = 0$, and $\mathrm{Var}(\xi_j^{(i)}) = w_j^2/m$. Let $s_m^2 = \sum_{j=1}^m \mathrm{Var}(\xi_j^{(i)}) = \|w\|_2^2/m$. By the Lindeberg--Feller theorem~\cite{billingsley1986probability}, it suffices to check
\[
\sum_{j=1}^m
\mathbb{E}\!\left[
\frac{(\xi_j^{(i)})^2}{s_m^2}
\mathbf{1}_{\{|\xi_j^{(i)}| > \varepsilon s_m\}}
\right] \to 0.
\]
Since $|\xi_j^{(i)}| = |w_j|/\sqrt{m}$ and $s_m = \|w\|_2/\sqrt{m}$, the indicator is equivalent to $|w_j| > \varepsilon \|w\|_2$, and
\[
\frac{(\xi_j^{(i)})^2}{s_m^2} = \frac{w_j^2}{\|w\|_2^2}.
\]
Thus, the sum reduces to
\[
\sum_{j=1}^m
\frac{w_j^2}{\|w\|_2^2}
\mathbf{1}_{\{|w_j| > \varepsilon \|w\|_2\}},
\]
which is zero for all sufficiently large $m$ by the assumption $\|w\|_\infty/\|w\|_2 \to 0$. Hence the claim follows.
\hfill $\blacksquare$

\begin{figure}[h!]
    \centering
    \includegraphics[width=0.48\linewidth]{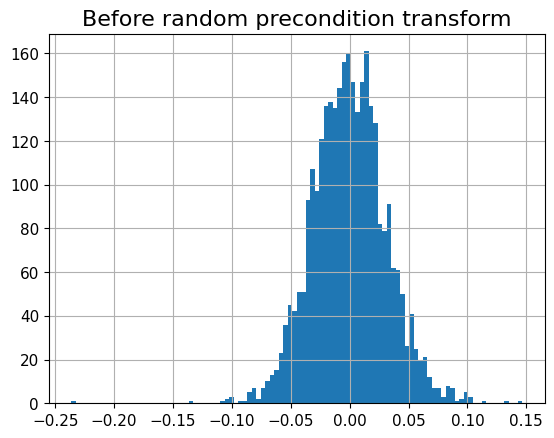}
    \hfill
    \includegraphics[width=0.48\linewidth]{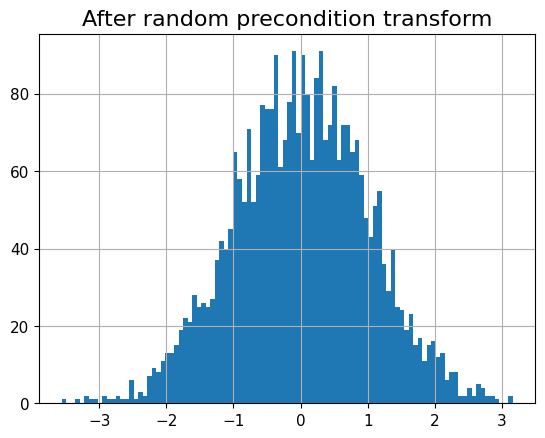}
    \caption{Empirical distributions of the coordinates of the first column vector of the query projection matrix in layer 0 of Qwen3-0.6B, before and after random Hadamard preconditioning. After preconditioning, the distribution becomes noticeably closer to Gaussian, which is consistent with the asymptotic normality stated in the lemma.}
    \label{fig:rht_histograms}
\end{figure}

Figure~\ref{fig:rht_histograms} illustrates the lemma empirically on the first column vector of the query projection matrix in layer 0 of Qwen3-0.6B.

    \section{Additional Results}\label{appendix:additional_results}

\subsection{Qwen3 models}

We evaluate Qwen3 models on zero-shot benchmarks, including BoolQ~\cite{clark2019boolq}, WinoGrande~\cite{sakaguchi2019winogrande}, ARC-Easy~\cite{Clark2018ThinkYH}, ARC-Challenge~\cite{Clark2018ThinkYH}, and PIQA~\cite{Bisk2020}, we report five-shot accuracy on MMLU~\cite{hendryckstest2021, hendrycks2021ethics}. We refer to~\citet{zheng2025empiricalstudyqwen3quantization}.

\begin{center}
\captionsetup{type=table}
\captionof{table}{Zero-shot accuracy ($\uparrow$) comparison on Qwen3-0.6B, Qwen3-1.7B, Qwen3-4B, and Qwen3-8B under low-bit quantization. BoolQ, WinoGrande, ARC-Easy, ARC-Challenge, and PIQA are evaluated in the zero-shot setting, while MMLU is evaluated in the five-shot setting. Gray rows denote our method. All results are reported using LM Eval Harness v0.4.11.}
\label{tab:qwen_main}
\scriptsize
\setlength{\tabcolsep}{4pt}
\begin{adjustbox}{width=\textwidth}
\begin{tabular}{llccccccc}
\toprule
\textbf{Model} & \textbf{Method} & \textbf{Bits} & \textbf{BoolQ}$\uparrow$ & \textbf{WinoGrande}$\uparrow$ & \textbf{ArcE}$\uparrow$ & \textbf{ArcC}$\uparrow$ & \textbf{PiQA}$\uparrow$ & \textbf{MMLU}$\uparrow$ \\
\midrule

\multirow{8}{*}{Qwen3-0.6B}
& FP16  & 16    & 64.31 & 55.96 & 60.82 & 31.40 & 67.74 & 47.00 \\
& AWQ   & 3     & \textbf{57.50} & 50.90 & 38.60 & 21.60 & 60.00 & 26.40 \\
& \cellcolor{gray!15} HADVQ & \cellcolor{gray!15} 3.41  & \cellcolor{gray!15}53.15 & \cellcolor{gray!15}\textbf{54.30} & \cellcolor{gray!15}\textbf{58.08} & \cellcolor{gray!15}\textbf{28.33} & \cellcolor{gray!15}\textbf{65.29} & \cellcolor{gray!15}\textbf{41.40} \\
& RTN   & 2     & 45.80 & 51.50 & 23.70 & 22.50 & 51.50 & 24.40 \\
& AWQ   & 2     & 50.40 & 51.90 & 23.60 & 22.30 & 52.70 & 25.50 \\
& GPTQ  & 2     & 44.30 & 49.50 & 25.20 & 21.40 & 51.90 & 24.50 \\
& \cellcolor{gray!15} HADVQ & \cellcolor{gray!15} 2.535 & \cellcolor{gray!15}\textbf{60.34} & \cellcolor{gray!15}\textbf{55.33} & \cellcolor{gray!15}\textbf{40.70} & \cellcolor{gray!15}\textbf{23.63} & \cellcolor{gray!15}\textbf{60.61} & \cellcolor{gray!15}\textbf{29.80} \\
& \cellcolor{gray!15} HADVQ & \cellcolor{gray!15} 2.41  & \cellcolor{gray!15}56.12 & \cellcolor{gray!15}50.99 & \cellcolor{gray!15}40.40 & \cellcolor{gray!15}22.18 & \cellcolor{gray!15}58.98 & \cellcolor{gray!15}28.80 \\
\midrule

\multirow{8}{*}{Qwen3-1.7B}
& FP16  & 16    & 77.60 & 60.90 & 72.40 & 40.10 & 72.50 & 60.00 \\
& AWQ   & 3     & 61.20 & 52.60 & 49.80 & 24.70 & 63.50 & 35.60 \\
& \cellcolor{gray!15} HADVQ & \cellcolor{gray!15} 3.41  & \cellcolor{gray!15}\textbf{71.62} & \cellcolor{gray!15}\textbf{58.17} & \cellcolor{gray!15}\textbf{66.04} & \cellcolor{gray!15}\textbf{37.20} & \cellcolor{gray!15}\textbf{71.00} & \cellcolor{gray!15}\textbf{55.80} \\
& RTN   & 2     & 48.40 & 51.90 & 24.50 & 22.80 & 52.30 & 24.40 \\
& AWQ   & 2     & 45.00 & 50.40 & 25.00 & 22.40 & 54.80 & 24.20 \\
& GPTQ  & 2     & 44.50 & 50.70 & 25.30 & 21.90 & 53.70 & 24.40 \\
& \cellcolor{gray!15} HADVQ & \cellcolor{gray!15} 2.535 & \cellcolor{gray!15}67.00 & \cellcolor{gray!15}\textbf{56.27} & \cellcolor{gray!15}\textbf{59.09} & \cellcolor{gray!15}\textbf{30.29} & \cellcolor{gray!15}\textbf{66.81} & \cellcolor{gray!15}\textbf{43.40} \\
& \cellcolor{gray!15} HADVQ & \cellcolor{gray!15} 2.41  & \cellcolor{gray!15}\textbf{68.65} & \cellcolor{gray!15}53.91 & \cellcolor{gray!15}57.20 & \cellcolor{gray!15}27.99 & \cellcolor{gray!15}64.20 & \cellcolor{gray!15}41.00 \\
\midrule

\multirow{8}{*}{Qwen3-4B}
& FP16  & 16    & 85.10 & 65.80 & 80.50 & 50.60 & 75.00 & 69.70 \\
& AWQ   & 3     & 63.60 & 54.60 & 61.70 & 32.60 & 66.40 & 43.90 \\
& \cellcolor{gray!15} HADVQ & \cellcolor{gray!15} 3.41  & \cellcolor{gray!15}\textbf{84.10} & \cellcolor{gray!15}\textbf{64.80} & \cellcolor{gray!15}\textbf{77.19} & \cellcolor{gray!15}\textbf{48.81} & \cellcolor{gray!15}\textbf{73.39} & \cellcolor{gray!15}\textbf{67.00} \\
& RTN   & 2     & 48.90 & 49.90 & 24.60 & 22.00 & 52.00 & 24.00 \\
& AWQ   & 2     & 47.70 & 51.90 & 23.40 & 22.10 & 49.90 & 24.70 \\
& GPTQ  & 2     & 41.80 & 50.40 & \textbf{80.10} & 21.30 & 52.90 & 24.60 \\
& \cellcolor{gray!15} HADVQ & \cellcolor{gray!15} 2.535 & \cellcolor{gray!15}79.27 & \cellcolor{gray!15}\textbf{58.48} & \cellcolor{gray!15}73.65 & \cellcolor{gray!15}\textbf{41.55} & \cellcolor{gray!15}71.16 & \cellcolor{gray!15}\textbf{59.50} \\
& \cellcolor{gray!15} HADVQ & \cellcolor{gray!15} 2.41  & \cellcolor{gray!15}\textbf{79.27} & \cellcolor{gray!15}58.41 & \cellcolor{gray!15}70.88 & \cellcolor{gray!15}40.36 & \cellcolor{gray!15}\textbf{71.55} & \cellcolor{gray!15}57.40 \\
\midrule

\multirow{8}{*}{Qwen3-8B}
& FP16  & 16    & 86.50 & 68.00 & 83.50 & 55.50 & 76.40 & 74.70 \\
& AWQ   & 3     & 78.50 & 62.20 & 68.20 & 41.00 & 71.90 & 52.20 \\
& \cellcolor{gray!15} HADVQ & \cellcolor{gray!15} 3.41  & \cellcolor{gray!15}\textbf{85.57} & \cellcolor{gray!15}\textbf{68.35} & \cellcolor{gray!15}\textbf{80.81} & \cellcolor{gray!15}\textbf{50.34} & \cellcolor{gray!15}\textbf{76.99} & \cellcolor{gray!15}\textbf{72.40} \\
& RTN   & 2     & 49.80 & 51.20 & 25.30 & 22.30 & 52.90 & 24.40 \\
& AWQ   & 2     & 46.60 & 48.80 & 25.30 & 21.40 & 52.60 & 24.70 \\
& GPTQ  & 2     & 41.80 & 49.30 & 25.80 & 22.80 & 51.80 & 24.20 \\
& \cellcolor{gray!15} HADVQ & \cellcolor{gray!15} 2.535 & \cellcolor{gray!15}\textbf{84.92} & \cellcolor{gray!15}63.77 & \cellcolor{gray!15}\textbf{79.46} & \cellcolor{gray!15}\textbf{49.49} & \cellcolor{gray!15}73.83 & \cellcolor{gray!15}\textbf{66.90} \\
& \cellcolor{gray!15} HADVQ & \cellcolor{gray!15} 2.41  & \cellcolor{gray!15}82.23 & \cellcolor{gray!15}\textbf{65.98} & \cellcolor{gray!15}78.45 & \cellcolor{gray!15}48.29 & \cellcolor{gray!15}\textbf{74.81} & \cellcolor{gray!15}66.40 \\
\bottomrule
\end{tabular}
\end{adjustbox}
\end{center}

We also evaluate on token perplexity benchmarks, including WikiText-2~\cite{merity2016pointer} and a 256-sample subset of C4~\cite{raffel2023exploringlimitstransferlearning} with a sequence length of 2048.

\begin{center}
\captionsetup{type=table}
\captionof{table}{Token perplexity ($\downarrow$) on WikiText-2 and C4 for Qwen3 models under low-bit quantization.}
\label{tab:qwen_ppl}
\scriptsize
\setlength{\tabcolsep}{3.2pt}
\renewcommand{\arraystretch}{0.95}
\begin{adjustbox}{width=\textwidth}
\begin{tabular}{l c cc cc cc cc}
\toprule
& & \multicolumn{2}{c}{Qwen3-0.6B} & \multicolumn{2}{c}{Qwen3-1.7B} & \multicolumn{2}{c}{Qwen3-4B} & \multicolumn{2}{c}{Qwen3-8B} \\
\cmidrule(lr){3-4} \cmidrule(lr){5-6} \cmidrule(lr){7-8} \cmidrule(lr){9-10}
\textbf{Method} & \textbf{bits} & \textbf{Wiki2}$\downarrow$ & \textbf{C4}$\downarrow$ & \textbf{Wiki2}$\downarrow$ & \textbf{C4}$\downarrow$ & \textbf{Wiki2}$\downarrow$ & \textbf{C4}$\downarrow$ & \textbf{Wiki2}$\downarrow$ & \textbf{C4}$\downarrow$ \\
\midrule

FP16  & 16   & 20.90 & 25.40 & 16.70 & 19.20 & 13.70 & 16.60 & 9.71 & 13.30 \\
AWQ   & 3    & 80.40 & 72.90 & 29.00 & 32.50 & 33.40 & 29.90 & 14.90 & 18.50 \\
\rowcolor{gray!15}
HADVQ & 3.41 & \textbf{24.57} & \textbf{28.77} & \textbf{17.55} & \textbf{20.59} & \textbf{14.24} & \textbf{17.41} & \textbf{10.18} & \textbf{13.89} \\
RTN   & 2     & 9.090E7 & 5.66E7 & 1.65E7 & 1.63E7 & 9.31E6 & 8.17E6 & NaN & NaN \\
AWQ   & 2     & 3.86E7 & 3.43E7 & 6.71E6 & 7.97E6 & 6.53E6 & 5.55E6 & 4.16E7 & 2.89E7 \\
GPTQ  & 2     & 7.19E5 & 5.91E5 & 1.43E5 & 3.89E3 & 2.06E4 & 1.25E4 & 4.66E3 & 2.93E3 \\
\rowcolor{gray!15}
HADVQ & 2.535 & \textbf{41.92} & \textbf{44.68} & \textbf{22.11} & \textbf{25.69} & \textbf{17.03} & \textbf{19.69} & \textbf{11.58} & \textbf{15.43} \\
\rowcolor{gray!15}
HADVQ & 2.41  & 49.39 & 54.32 & 25.90 & 28.46 & 17.73 & 20.89 & 12.10 & 15.84 \\
\bottomrule
\end{tabular}
\end{adjustbox}
\end{center}

\subsection{Llama models results for different LM Eval Harness versions}

\paragraph{LM Eval Harness v0.3.0.}

Addition to the table~\ref{tab:llama_main} we also report results on Llama3.2-1B and Llama3.2-3B. For all benchmarks, we report accuracy (\texttt{acc}).

\begin{center}
\captionsetup{type=table}
\captionof{table}{Zero-shot accuracy ($\uparrow$) comparison on Llama3.2-1/3B models using LM Eval Harness v0.4.11. We report \texttt{acc} for all benchmarks.}
\label{tab:llama32_report_acc}
\scriptsize
\setlength{\tabcolsep}{3pt}
\renewcommand{\arraystretch}{0.93}
\begin{adjustbox}{width=\textwidth}
\begin{tabular}{llccccccc}
\toprule
\textbf{Model} & \textbf{Method} & \textbf{Bits} & \textbf{BoolQ}$\uparrow$ & \textbf{Wino}$\uparrow$ & \textbf{ArcE}$\uparrow$ & \textbf{ArcC}$\uparrow$ & \textbf{Hella}$\uparrow$ & \textbf{PiQA}$\uparrow$ \\
\midrule

\multirow{4}{*}{Llama3.2-1B}
& FP16   & 16    &  64.46 & 60.62 & 65.49 & 31.48 & 47.76 & 74.43 \\
& \cellcolor{gray!15} HADVQ & \cellcolor{gray!15} 2.41 &  \cellcolor{gray!15}61.5 & \cellcolor{gray!15}52.88 & \cellcolor{gray!15}45.33 & \cellcolor{gray!15}23.12 & \cellcolor{gray!15}33.58 & \cellcolor{gray!15}61.75 \\
& \cellcolor{gray!15} HADVQ & \cellcolor{gray!15} 2.535 &  \cellcolor{gray!15}60.58 & \cellcolor{gray!15}52.88 & \cellcolor{gray!15}48.95 & \cellcolor{gray!15}23.46 & \cellcolor{gray!15}36.01 & \cellcolor{gray!15}65.51 \\
& \cellcolor{gray!15} HADVQ & \cellcolor{gray!15} 3.41 &  \cellcolor{gray!15}63.52 & \cellcolor{gray!15}58.48 & \cellcolor{gray!15}57.66 & \cellcolor{gray!15}27.65 & \cellcolor{gray!15}41.46 & \cellcolor{gray!15}71.27 \\
\midrule

\multirow{4}{*}{Llama3.2-3B}
& FP16   & 16 & 73.55 & 69.77 & 74.62 & 42.24 & 55.24 & 76.71 \\
& \cellcolor{gray!15} HADVQ & \cellcolor{gray!15} 2.41 & \cellcolor{gray!15}50.18 & \cellcolor{gray!15}61.64 & \cellcolor{gray!15}61.83 & \cellcolor{gray!15}32.00 & \cellcolor{gray!15}43.88 & \cellcolor{gray!15}72.20 \\
& \cellcolor{gray!15} HADVQ & \cellcolor{gray!15} 2.535 & \cellcolor{gray!15}69.33 & \cellcolor{gray!15}63.61 & \cellcolor{gray!15}64.10 & \cellcolor{gray!15}32.85 & \cellcolor{gray!15}46.51 & \cellcolor{gray!15}73.01 \\
& \cellcolor{gray!15} HADVQ & \cellcolor{gray!15} 3.41 & \cellcolor{gray!15}65.47 & \cellcolor{gray!15}67.48 & \cellcolor{gray!15}69.23 & \cellcolor{gray!15}39.16 & \cellcolor{gray!15}51.15 & \cellcolor{gray!15}75.57 \\

\bottomrule
\end{tabular}
\end{adjustbox}
\end{center}

\begin{center}
\captionsetup{type=table}
\captionof{table}{Token perplexity ($\downarrow$) on WikiText-2 with 2048 and 4096 sequences lengths.}
\label{tab:wikitext_ppl_by_sequence_length}
\scriptsize
\setlength{\tabcolsep}{3.2pt}
\renewcommand{\arraystretch}{0.95}
\begin{adjustbox}{width=0.6\textwidth}
\begin{tabular}{l c cc cc}
\toprule
& & \multicolumn{2}{c}{Llama3.2-1B} & \multicolumn{2}{c}{Llama3.2-3B} \\
\cmidrule(lr){3-4} \cmidrule(lr){5-6}
\textbf{Method} & \textbf{bits} & \textbf{2048}$\downarrow$ & \textbf{4096}$\downarrow$ & \textbf{2048}$\downarrow$ & \textbf{4096}$\downarrow$ \\
\midrule

FP16  & 16   & 9.75	& 9.06 & 7.81 & 7.28 \\
\rowcolor{gray!15}
HADVQ & 3.41 & 11.98 & 11.09 & 8.69 & 8.1 \\
\rowcolor{gray!15}
HADVQ & 2.535 & 22.1 & 20.8 & 11.67 & 10.91 \\
\rowcolor{gray!15}
HADVQ & 2.41  & 29.1 & 27.57 & 13.13 & 12.24 \\
\bottomrule
\end{tabular}
\end{adjustbox}
\end{center}

\paragraph{LM Eval Harness v0.4.4.}
We also report results obtained with LM Eval Harness v0.4.4 on models from the Llama family~\cite{llama}, including Llama2-7B, Llama2-13B, Llama3-8B, Llama3.2-1B, and Llama3.2-3B. Compared with the v0.3.0 setup used in the main text, this evaluation additionally includes recent scalar PTQ baselines such as ReSpinQuant~\cite{kim2026respinquantefficientlayerwisellm}, QuaRot~\cite{ashkboos2024quarot}, and OSTQuant~\cite{hu2025ostquant}. For BoolQ~\cite{clark2019boolq}, WinoGrande~\cite{sakaguchi2019winogrande}, ARC-Easy~\cite{Clark2018ThinkYH}, ARC-Challenge~\cite{Clark2018ThinkYH} and PIQA~\cite{Bisk2020}, we report the normalized scores.

\begin{center}
\captionsetup{type=table}
\caption{Zero-shot accuracy ($\uparrow$) comparison on Llama-family models using LM Eval Harness v0.4.4. For ARC-Easy, ARC-Challenge, HellaSwag, and PIQA, we report normalized scores.}
\label{tab:llama_lmeval_044}
\scriptsize
\setlength{\tabcolsep}{3pt}
\renewcommand{\arraystretch}{0.93}
\begin{adjustbox}{width=\textwidth}
\begin{tabular}{llccccccc}
\toprule
\textbf{Model} & \textbf{Method} & \textbf{Bits} & \textbf{BoolQ}$\uparrow$ & \textbf{Wino}$\uparrow$ & \textbf{ArcE}$\uparrow$ & \textbf{ArcC}$\uparrow$ & \textbf{Hella}$\uparrow$ & \textbf{PiQA}$\uparrow$ \\
\midrule

\multirow{10}{*}{Llama2-7B}
& FP16        & 16   & 77.74 & 68.82 & 74.58 & 46.16 & 75.99 & 79.11 \\

& ReSpinQuant & 4    & 74.89 & 66.22 & 70.20 & 43.94 & 73.38 & 76.17  \\
& QuaRot      & 4    & 74.13 & 66.85 & 70.24 & 43.52 & 73.27 & 77.09  \\
& OSTQuant    & 4    & 74.31 & 66.85 & 69.23 & 41.72 & 73.78 & 76.88  \\

& ReSpinQuant & 3    & 65.84 & 55.72 & 49.41 & 29.78 & 59.52 & 67.95 \\
& QuaRot      & 3    & 56.36 & 48.38 & 43.22 & 27.13 & 41.73 & 41.73 \\
& OSTQuant    & 3    & 60.12 & 55.56 & 52.23 & 29.95 & 56.14 & 68.88 \\

& \cellcolor{gray!15} HADVQ & \cellcolor{gray!15} 3.41  & \cellcolor{gray!15}\textbf{76.42} & \cellcolor{gray!15}\textbf{68.35} & \cellcolor{gray!15}\textbf{74.16} & \cellcolor{gray!15}\textbf{45.56} & \cellcolor{gray!15}\textbf{74.33} & \cellcolor{gray!15}\textbf{78.45} \\

& \cellcolor{gray!15} HADVQ & \cellcolor{gray!15} 2.535 & \cellcolor{gray!15}72.94 & \cellcolor{gray!15}67.96 & \cellcolor{gray!15}69.78 & \cellcolor{gray!15}40.53 & \cellcolor{gray!15}71.24 & \cellcolor{gray!15}76.71 \\

& \cellcolor{gray!15} HADVQ & \cellcolor{gray!15} 2.41 & \cellcolor{gray!15}74.07
& \cellcolor{gray!15}68.19	& \cellcolor{gray!15}68.43 & \cellcolor{gray!15}42.06 & \cellcolor{gray!15}70.55 & \cellcolor{gray!15}76.93 \\

\midrule

\multirow{10}{*}{Llama2-13B}
& FP16        & 16   & 80.52 & 72.20 & 77.48 & 49.23 & 79.39 & 80.52  \\

& ReSpinQuant & 4    & 79.42 & 69.77 & 74.87 & 46.76 & 77.24 & 78.94  \\
& QuaRot      & 4    & 79.27 & 69.53 & 75.46 & 45.82 & 76.19 & 78.18  \\
& OSTQuant    & 4    & 78.62 & 69.14 & 73.36 & 48.29 & 77.04 & 78.89  \\

& ReSpinQuant & 3    & 70.46 & 59.59 & 60.23 & 36.77 & 67.20 & 72.42  \\
& QuaRot      & 3    & 58.35 & 51.54 & 47.69 & 27.30 & 45.84 & 64.04  \\
& OSTQuant    & 3    & 69.27 & 61.09 & 59.22 & 34.90 & 66.48 & 72.31  \\

& \cellcolor{gray!15} HADVQ & \cellcolor{gray!15} 3.41  & \cellcolor{gray!15}79.54 & \cellcolor{gray!15}\textbf{72.93} & \cellcolor{gray!15}\textbf{76.64} & \cellcolor{gray!15}\textbf{48.38} & \cellcolor{gray!15}\textbf{78.34} & \cellcolor{gray!15}\textbf{80.03}  \\

& \cellcolor{gray!15} HADVQ & \cellcolor{gray!15} 2.535 & \cellcolor{gray!15}\textbf{80.21} & \cellcolor{gray!15}70.17 & \cellcolor{gray!15}72.22 & \cellcolor{gray!15}45.56 & \cellcolor{gray!15}76.05 & \cellcolor{gray!15}78.70  \\

& \cellcolor{gray!15} HADVQ & \cellcolor{gray!15} 2.41 & \cellcolor{gray!15}78.96	& \cellcolor{gray!15}70.72 & 
\cellcolor{gray!15}71.51 & \cellcolor{gray!15}43.52 & \cellcolor{gray!15}75.05 & \cellcolor{gray!15}78.84 \\

\midrule

\multirow{10}{*}{Llama3-8B}
& FP16        & 16   & 81.35 & 72.45 & 77.78 & 53.33 & 79.14 & 80.85  \\

& ReSpinQuant & 4    & 77.49 & 68.11 & 75.84 & 47.10 & 75.55 & \textbf{79.11}  \\
& QuaRot      & 4    & 76.61 & 67.48 & 67.59 & 44.80 & 74.83 & 77.58  \\
& OSTQuant    & 4    & 77.83 & 69.38 & 75.55 & 47.10 & 74.44 & 77.15  \\

& ReSpinQuant & 3    & 65.75 & 55.88 & 54.63 & 31.91 & 59.86 & 67.41  \\
& QuaRot      & 3    & 50.34 & 51.30 & 31.23 & 22.53 & 31.68 & 52.94  \\
& OSTQuant    & 3    & 60.15 & 55.01 & 47.31 & 29.78 & 56.04 & 65.13  \\

& \cellcolor{gray!15} HADVQ & \cellcolor{gray!15} 3.41  & \cellcolor{gray!15}\textbf{80.98} & \cellcolor{gray!15}\textbf{74.19} & \cellcolor{gray!15}\textbf{77.40} & \cellcolor{gray!15}\textbf{52.05} & \cellcolor{gray!15}\textbf{76.66} & \cellcolor{gray!15}78.84  \\

& \cellcolor{gray!15} HADVQ & \cellcolor{gray!15} 2.535 & \cellcolor{gray!15}74.46 & \cellcolor{gray!15}70.56 & \cellcolor{gray!15}70.66 & \cellcolor{gray!15}44.37 & \cellcolor{gray!15}71.74 & \cellcolor{gray!15}76.66  \\

& \cellcolor{gray!15} HADVQ & \cellcolor{gray!15} 2.41 & \cellcolor{gray!15}73.58 &	\cellcolor{gray!15}67.72  & 
\cellcolor{gray!15}66.62 & \cellcolor{gray!15}39.59 & \cellcolor{gray!15}68.46 & \cellcolor{gray!15}75.41 \\

\midrule

\multirow{10}{*}{Llama3.2-1B}
& FP16        & 16   & 63.94 & 60.46 & 60.48 & 36.35 & 63.66 & 74.54  \\

& ReSpinQuant & 4    & 58.04 & 57.38 & 52.65 & 31.91 & 55.17 & 69.26  \\
& QuaRot      & 4    & 50.83 & 52.96 & 50.25 & 30.55 & 54.23 & 67.41  \\
& OSTQuant    & 4    & 57.77 & 53.59 & 52.02 & 31.14 & 55.71 & 69.37  \\

& ReSpinQuant & 3    & 56.48 & 50.67 & 36.74 & 21.76 & 35.04 & 58.00  \\
& QuaRot      & 3    & 49.94 & 49.57 & 27.99 & 23.29 & 27.84 & 50.65  \\
& OSTQuant    & 3    & 53.52 & 50.04 & 36.20 & 24.15 & 32.46 & 55.28  \\

& \cellcolor{gray!15} HADVQ & \cellcolor{gray!15} 3.41  & \cellcolor{gray!15}\textbf{64.01} & \cellcolor{gray!15}\textbf{59.43} & 
\cellcolor{gray!15}\textbf{56.90} & \cellcolor{gray!15}\textbf{33.79} & \cellcolor{gray!15}\textbf{59.69} & \cellcolor{gray!15}\textbf{72.63}  \\

& \cellcolor{gray!15} HADVQ & \cellcolor{gray!15} 2.535 & \cellcolor{gray!15}60.24 & \cellcolor{gray!15}53.75 & \cellcolor{gray!15}46.04 & \cellcolor{gray!15}28.50 & \cellcolor{gray!15}47.83 & \cellcolor{gray!15}67.30  \\

& \cellcolor{gray!15} HADVQ & \cellcolor{gray!15} 2.41 & \cellcolor{gray!15}62.2 & \cellcolor{gray!15}55.64 &
\cellcolor{gray!15}44.49 & \cellcolor{gray!15}27.30 & \cellcolor{gray!15}43.34 & \cellcolor{gray!15}62.68 \\

\midrule

\multirow{10}{*}{Llama3.2-3B}
& FP16        & 16   & 73.33 & 69.77 & 71.68 & 45.90 & 73.63 & 77.42  \\

& ReSpinQuant & 4    & 69.88 & 63.54 & 66.20 & 40.53 & 68.64 & 75.95  \\
& QuaRot      & 4    & 66.42 & 63.06 & 63.64 & 38.99 & 67.67 & 73.12  \\
& OSTQuant    & 4    & 66.39 & 63.06 & 66.37 & 41.47 & 69.31 & 74.70  \\

& ReSpinQuant & 3    & 61.47 & 52.09 & 45.12 & 27.99 & 48.31 & 60.88  \\
& QuaRot      & 3    & 44.43 & 51.38 & 29.46 & 22.10 & 30.67 & 55.11  \\
& OSTQuant    & 3    & 53.43 & 52.64 & 42.00 & 26.11 & 42.80 & 61.21  \\

& \cellcolor{gray!15} HADVQ & \cellcolor{gray!15} 3.41  & \cellcolor{gray!15}69.33 & \cellcolor{gray!15}\textbf{67.96} & \cellcolor{gray!15}\textbf{66.50} & \cellcolor{gray!15}\textbf{42.49} & \cellcolor{gray!15}\textbf{70.79} & \cellcolor{gray!15}\textbf{76.88}  \\

& \cellcolor{gray!15} HADVQ & \cellcolor{gray!15} 2.535 & \cellcolor{gray!15}\textbf{71.04} & \cellcolor{gray!15}63.93 & \cellcolor{gray!15}60.10 & \cellcolor{gray!15}36.01 & \cellcolor{gray!15}64.22 & \cellcolor{gray!15}73.34  \\

& \cellcolor{gray!15} HADVQ & \cellcolor{gray!15} 2.41 & \cellcolor{gray!15}55.11 & \cellcolor{gray!15}62.35 & \cellcolor{gray!15}61.20 & \cellcolor{gray!15}36.86 & \cellcolor{gray!15}60.92 & \cellcolor{gray!15}72.69 \\

\bottomrule
\end{tabular}
\end{adjustbox}
\end{center}

In addition, we report MMLU~\cite{hendryckstest2021, hendrycks2021ethics} with zero-shot and five-shot settings.

\begin{center}
\captionsetup{type=table}
\captionof{table}{MMLU accuracy ($\uparrow$) with zero-shot and five-shot settings. All results are reported using LM Eval Harness v0.4.4.}
\label{tab:mmlu_shot_comparison}
\scriptsize
\setlength{\tabcolsep}{3.2pt}
\renewcommand{\arraystretch}{0.95}
\begin{adjustbox}{width=\textwidth}
\begin{tabular}{l c cc cc cc cc cc}
\toprule
& & \multicolumn{2}{c}{Llama2-7B} & \multicolumn{2}{c}{Llama2-13B} & \multicolumn{2}{c}{Llama3-8B} & \multicolumn{2}{c}{Llama3.2-1B} & \multicolumn{2}{c}{Llama3.2-3B} \\
\cmidrule(lr){3-4} \cmidrule(lr){5-6} \cmidrule(lr){7-8} \cmidrule(lr){9-10} \cmidrule(lr){11-12}
\textbf{Method} & \textbf{bits} & \textbf{0-shot}$\uparrow$ & \textbf{5-shot}$\uparrow$ & \textbf{0-shot}$\uparrow$ & \textbf{5-shot}$\uparrow$ & \textbf{0-shot}$\uparrow$ & \textbf{5-shot}$\uparrow$ & \textbf{0-shot}$\uparrow$ & \textbf{5-shot}$\uparrow$ & \textbf{0-shot}$\uparrow$ & \textbf{5-shot}$\uparrow$ \\
\midrule

FP16  & 16   & 41.84 & 45.84 & 52.07 & 55.17 & 62.15 & 65.33 & 37.68 & 32. & 54.37 & 56.38 \\

\rowcolor{gray!15}
HADVQ & 3.41 & 31.09 & 35.19 & 44.49 & 48.29 & 47.51 &	47.40 & 25.34 &	24.43 & 37.37 &	37.32 \\ 
\rowcolor{gray!15}
HADVQ & 2.535 & 31.44 & 40.34 & 48.00 & 49.32 & 49.76 &	52.25 & 25.35 &	26.83 & 38.67 &	41.63 \\
\rowcolor{gray!15}
HADVQ & 2.41 & 36.92 & 44.43 & 50.64 & 53.07 & 59.02 &	61.48 & 30.78 &	26.83 & 49.12 &	52.19 \\

\bottomrule
\end{tabular}
\end{adjustbox}
\end{center}

\paragraph{LM Eval Harness v0.4.11.}

We also report results obtained with LM Eval Harness v0.4.11 on models from the Llama family~\cite{llama}, including Llama2-7B, Llama2-13B, Llama3-8B, Llama3.2-1B, and Llama3.2-3B. For all benchmarks, we report accuracy (\texttt{acc}).

\begin{center}
\captionsetup{type=table}
\captionof{table}{Zero-shot accuracy ($\uparrow$) comparison on Llama-family models using LM Eval Harness v0.4.11. We report \texttt{acc} for all benchmarks.}
\label{tab:llama_lmeval_0411}
\scriptsize
\setlength{\tabcolsep}{3pt}
\renewcommand{\arraystretch}{0.93}
\begin{adjustbox}{width=\textwidth}
\begin{tabular}{llccccccc}
\toprule
\textbf{Model} & \textbf{Method} & \textbf{Bits} & \textbf{BoolQ}$\uparrow$ & \textbf{Wino}$\uparrow$ & \textbf{ArcE}$\uparrow$ & \textbf{ArcC}$\uparrow$ & \textbf{Hella}$\uparrow$ & \textbf{PiQA}$\uparrow$ \\
\midrule

\multirow{4}{*}{Llama2-7B}
& FP16   & 16    & 79.39 & 69.38 & 75.51 & 42.92 & 57.12 & 78.13 \\
& \cellcolor{gray!15} HADVQ & \cellcolor{gray!15} 2.41  & \cellcolor{gray!15}74.46 & \cellcolor{gray!15}67.56 & \cellcolor{gray!15}70.92 & \cellcolor{gray!15}38.82 & \cellcolor{gray!15}51.84 & \cellcolor{gray!15}76.50 \\
& \cellcolor{gray!15} HADVQ & \cellcolor{gray!15} 2.535 & \cellcolor{gray!15}74.74 & \cellcolor{gray!15}68.43 & \cellcolor{gray!15}73.48 & \cellcolor{gray!15}37.54 & \cellcolor{gray!15}52.76 & \cellcolor{gray!15}76.55 \\
& \cellcolor{gray!15} HADVQ & \cellcolor{gray!15} 3.41  & \cellcolor{gray!15}\textbf{78.38} & \cellcolor{gray!15}\textbf{68.51} & \cellcolor{gray!15}\textbf{75.04} & \cellcolor{gray!15}\textbf{42.06} & \cellcolor{gray!15}\textbf{55.97} & \cellcolor{gray!15}\textbf{77.37} \\
\midrule

\multirow{4}{*}{Llama2-13B}
& FP16   & 16    & 82.14 & 72.61 & 78.96 & 47.35 & 60.20 & 79.54 \\
& \cellcolor{gray!15} HADVQ & \cellcolor{gray!15} 2.41  & \cellcolor{gray!15}80.15 & \cellcolor{gray!15}70.40 & \cellcolor{gray!15}75.63 & \cellcolor{gray!15}41.89 & \cellcolor{gray!15}56.55 & \cellcolor{gray!15}77.37 \\
& \cellcolor{gray!15} HADVQ & \cellcolor{gray!15} 2.535 & \cellcolor{gray!15}81.44 & \cellcolor{gray!15}71.03 & \cellcolor{gray!15}75.51 & \cellcolor{gray!15}41.98 & \cellcolor{gray!15}57.12 & \cellcolor{gray!15}77.86 \\
& \cellcolor{gray!15} HADVQ & \cellcolor{gray!15} 3.41  & \cellcolor{gray!15}\textbf{81.62} & \cellcolor{gray!15}\textbf{72.61} & \cellcolor{gray!15}\textbf{78.66} & \cellcolor{gray!15}\textbf{45.22} & \cellcolor{gray!15}\textbf{59.55} & \cellcolor{gray!15}\textbf{79.16} \\
\midrule

\multirow{4}{*}{Llama3-8B}
& FP16   & 16    & 82.26 & 73.95 & 80.77 & 51.88 & 60.67 & 78.84 \\
& \cellcolor{gray!15} HADVQ & \cellcolor{gray!15} 2.41  & \cellcolor{gray!15}73.58 & \cellcolor{gray!15}67.01 & \cellcolor{gray!15}70.92 & \cellcolor{gray!15}37.29 & \cellcolor{gray!15}50.90 & \cellcolor{gray!15}74.37 \\
& \cellcolor{gray!15} HADVQ & \cellcolor{gray!15} 2.535 & \cellcolor{gray!15}75.26 & \cellcolor{gray!15}69.77 & \cellcolor{gray!15}73.11 & \cellcolor{gray!15}40.27 & \cellcolor{gray!15}53.44 & \cellcolor{gray!15}75.90 \\
& \cellcolor{gray!15} HADVQ & \cellcolor{gray!15} 3.41  & \cellcolor{gray!15}\textbf{81.59} & \cellcolor{gray!15}\textbf{73.40} & \cellcolor{gray!15}\textbf{79.34} & \cellcolor{gray!15}\textbf{47.35} & \cellcolor{gray!15}\textbf{58.04} & \cellcolor{gray!15}\textbf{78.40} \\
\midrule

\multirow{4}{*}{Llama3.2-1B}
& FP16   & 16    & 63.82 & 60.77 & 66.29 & 31.48 & 48.18 & 75.19 \\
& \cellcolor{gray!15} HADVQ & \cellcolor{gray!15} 2.41  & \cellcolor{gray!15}61.90 & \cellcolor{gray!15}54.30 & \cellcolor{gray!15}47.69 & \cellcolor{gray!15}25.09 & \cellcolor{gray!15}35.18 & \cellcolor{gray!15}65.02 \\
& \cellcolor{gray!15} HADVQ & \cellcolor{gray!15} 2.535 & \cellcolor{gray!15}60.98 & \cellcolor{gray!15}54.85 & \cellcolor{gray!15}51.47 & \cellcolor{gray!15}24.83 & \cellcolor{gray!15}37.81 & \cellcolor{gray!15}67.36 \\
& \cellcolor{gray!15} HADVQ & \cellcolor{gray!15} 3.41  & \cellcolor{gray!15}\textbf{63.70} & \cellcolor{gray!15}\textbf{59.04} & \cellcolor{gray!15}\textbf{62.79} & \cellcolor{gray!15}\textbf{28.67} & \cellcolor{gray!15}\textbf{44.82} & \cellcolor{gray!15}\textbf{73.67} \\
\midrule

\multirow{4}{*}{Llama3.2-3B}
& FP16   & 16    & 74.19 & 69.46 & 74.54 & 43.00 & 55.82 & 76.44 \\
& \cellcolor{gray!15} HADVQ & \cellcolor{gray!15} 2.41  & \cellcolor{gray!15}53.33 & \cellcolor{gray!15}62.04 & \cellcolor{gray!15}65.66 & \cellcolor{gray!15}34.81 & \cellcolor{gray!15}45.80 & \cellcolor{gray!15}72.85 \\
& \cellcolor{gray!15} HADVQ & \cellcolor{gray!15} 2.535 & \cellcolor{gray!15}\textbf{71.31} & \cellcolor{gray!15}65.59 & \cellcolor{gray!15}67.34 & \cellcolor{gray!15}34.64 & \cellcolor{gray!15}48.01 & \cellcolor{gray!15}73.67 \\
& \cellcolor{gray!15} HADVQ & \cellcolor{gray!15} 3.41  & \cellcolor{gray!15}71.22 & \cellcolor{gray!15}\textbf{68.35} & \cellcolor{gray!15}\textbf{72.81} & \cellcolor{gray!15}\textbf{41.21} & \cellcolor{gray!15}\textbf{53.20} & \cellcolor{gray!15}\textbf{76.44} \\
\bottomrule
\end{tabular}
\end{adjustbox}
\end{center}

In addition, we report MMLU~\cite{hendryckstest2021, hendrycks2021ethics} in both zero-shot and five-shot settings.

\begin{center}
\captionsetup{type=table}
\captionof{table}{MMLU accuracy ($\uparrow$) with zero-shot and five-shot settings. All results are reported using LM Eval Harness v0.4.11.}
\label{tab:llama_mmlu_0411}
\scriptsize
\setlength{\tabcolsep}{3.2pt}
\renewcommand{\arraystretch}{0.95}
\begin{adjustbox}{width=\textwidth}
\begin{tabular}{l c cc cc cc cc cc}
\toprule
& & \multicolumn{2}{c}{Llama2-7B} & \multicolumn{2}{c}{Llama2-13B} & \multicolumn{2}{c}{Llama3-8B} & \multicolumn{2}{c}{Llama3.2-1B} & \multicolumn{2}{c}{Llama3.2-3B} \\
\cmidrule(lr){3-4} \cmidrule(lr){5-6} \cmidrule(lr){7-8} \cmidrule(lr){9-10} \cmidrule(lr){11-12}
\textbf{Method} & \textbf{bits} & \textbf{0-shot}$\uparrow$ & \textbf{5-shot}$\uparrow$ & \textbf{0-shot}$\uparrow$ & \textbf{5-shot}$\uparrow$ & \textbf{0-shot}$\uparrow$ & \textbf{5-shot}$\uparrow$ & \textbf{0-shot}$\uparrow$ & \textbf{5-shot}$\uparrow$ & \textbf{0-shot}$\uparrow$ & \textbf{5-shot}$\uparrow$ \\
\midrule

FP16  & 16   & 41.71 & 45.98 & 52.36 & 55.50 & 62.41 & 65.48 & 38.48 & 32.17 & 55.13 & 56.68 \\
\rowcolor{gray!15}
HADVQ & 2.41  & 31.17 & 35.45 & 45.00 & 48.19 & 48.38 & 47.96 & 24.95 & 24.62 & 38.35 & 37.84 \\
\rowcolor{gray!15}
HADVQ & 2.535 & 32.83 & 40.42 & 47.91 & 49.52 & 50.37 & 52.68 & 25.94 & 27.30 & 42.33 & 41.71 \\
\rowcolor{gray!15}
HADVQ & 3.41  & 37.05 & 44.59 & 50.58 & 53.13 & 58.86 & 61.75 & 31.57 & 26.83 & 50.17 & 52.32 \\
\bottomrule
\end{tabular}
\end{adjustbox}
\end{center}
    \section{Broader Impacts}\label{appendix:impact}

This work studies post-training quantization for large language models. A potential positive impact is that improved ultra-low-bit quantization can reduce memory usage and deployment costs, which may make language models more accessible in resource-constrained settings and lower the hardware requirements needed for inference.
    \subsection{Inference Efficiency}\label{appendix:efficiency}

We also report a simple throughput comparison between full-precision and quantized models. Table~\ref{tab:throughput_qwen} shows generation speed in tokens per second measured on a single A100 40GB GPU.

At present, we optimized only the \texttt{pack\_int2} kernel for the magnitude codebook indices. Other parts of the quantized inference pipeline are not yet fully optimized. As a result, the reported throughput likely underestimates the efficiency that could be achieved with a more careful implementation.

\begin{table}[h!]
\centering
\caption{Inference throughput in tokens per second.}
\label{tab:throughput_qwen}
\small
\setlength{\tabcolsep}{8pt}
\renewcommand{\arraystretch}{1.05}
\begin{tabular}{lccc}
\toprule
\textbf{Bits} & \textbf{Qwen3-0.6B} & \textbf{Qwen3-4B} & \textbf{Qwen3-8B} \\
& \textbf{tok/s} & \textbf{tok/s} & \textbf{tok/s} \\
\midrule
16   & 21.48 & 17.05 & 16.73 \\
2.41 & 9.48  & 5.90  & 3.59  \\
\bottomrule
\end{tabular}
\end{table}
    \section{Licenses for Existing Assets}
\label{appendix:licenses}

We cite the main external assets used in this work in the paper and summarize their licenses below.

\begin{table}[H]
\centering
\caption{Main external assets used in this work and their licenses.}
\label{tab:asset_licenses}
\footnotesize
\setlength{\tabcolsep}{4pt}
\renewcommand{\arraystretch}{1.08}
\begin{tabularx}{\linewidth}{L{2.6cm} L{2.5cm} L{2.4cm} Y}
\toprule
\textbf{Asset} & \textbf{Type} & \textbf{License} & \textbf{Notes} \\
\midrule

Llama2, Llama3, Llama3.2
& Base models
& Llama Community License
& Use is subject to the corresponding Llama Community License Agreement and Acceptable Use Policy. \\
\midrule

Qwen3
& Base models
& Apache 2.0
& We use publicly released Qwen3 checkpoints. \\
\midrule

The Pile
& Calibration data
& MIT (repository)
& We use \texttt{mit-han-lab/pile-val-backup} for calibration. The Pile aggregates multiple sources, so users should also respect the terms of the underlying components. \\
\midrule

BoolQ
& Evaluation dataset
& CC BY-SA 3.0
& Used only for evaluation. \\
\midrule

WinoGrande
& Evaluation dataset
& Not clearly specified
& Used only for evaluation; we cite the original dataset source. \\
\midrule

ARC-Easy / ARC-Challenge
& Evaluation dataset
& Not clearly specified
& Used only for evaluation; we cite the original dataset source. \\
\midrule

HellaSwag
& Evaluation dataset
& MIT
& Used only for evaluation. \\
\midrule

PIQA
& Evaluation dataset
& Academic Free License v3.0
& Used only for evaluation. \\
\midrule

MMLU
& Evaluation dataset
& Not clearly specified
& Used only for evaluation; we cite the original dataset source. \\
\midrule

LM Eval Harness
& Evaluation framework
& MIT
& Used for downstream evaluation. \\
\midrule

PyTorch
& Software library
& BSD-style license
& Used for implementation. \\
\midrule

Transformers
& Software library
& Apache 2.0
& Used for model loading and inference. \\
\midrule

fast-hadamard-transform
& Software library
& BSD-3-Clause
& Used for randomized Hadamard preconditioning. \\
\midrule

torch\_kdtree
& Software library
& MIT
& Used for nearest-neighbor lookup in the directional codebook. \\
\bottomrule
\end{tabularx}
\end{table}
    
    \newpage
    \section*{NeurIPS Paper Checklist}

\begin{enumerate}

\item {\bf Claims}
    \item[] Question: Do the main claims made in the abstract and introduction accurately reflect the paper's contributions and scope?
    \item[] Answer: \answerYes{}
    \item[] Justification: The main claims in the abstract and introduction are consistent with the actual scope of the paper and are supported by both the experimental results and the ablation study. In particular, the paper claims that HADVQ combines random preconditioning, magnitude--direction decoupling, structured directional vector quantization, and Hessian-aware channel preservation, and these components are described in Section~3 and evaluated in Sections~4.2 and~4.3. The abstract and introduction also focus on the extremely low-bit post-training setting, which matches the experiments reported in the paper.

\item {\bf Limitations}
    \item[] Question: Does the paper discuss the limitations of the work performed by the authors?
    \item[] Answer: \answerYes{} 
    \item[] Justification: The paper includes a dedicated \textbf{Limitations} section discussing the main constraints of the current study, including the focus on weight-only PTQ, the fact that the method is not evaluated together with fine-tuning, the dependence of the directional codebook on 8-dimensional $E_8$ structure, and the additional runtime overhead introduced by the random Hadamard transform. It also clarifies that the extra memory used by the codebooks and the sensitive-channel mask is negligible when amortized over a model with billions of parameters.

\item {\bf Theory assumptions and proofs}
    \item[] Question: For each theoretical result, does the paper provide the full set of assumptions and a complete (and correct) proof?
    \item[] Answer: \answerYes{} 
    \item[] Justification: The paper includes one explicit theoretical result, stated as a lemma in Appendix~\ref{appendix:rht_prove}, together with its assumptions and a complete proof. In particular, the lemma states the asymptotic normality of coordinates after randomized Hadamard preconditioning under the assumption $\|w\|_\infty / \|w\|_2 \to 0$, and the proof is given using the Lindeberg--Feller theorem. The assumptions are stated directly in the lemma, and the proof is provided in full in the appendix.

    \item {\bf Experimental result reproducibility}
    \item[] Question: Does the paper fully disclose all the information needed to reproduce the main experimental results of the paper to the extent that it affects the main claims and/or conclusions of the paper (regardless of whether the code and data are provided or not)?
    \item[] Answer: \answerYes{} 
    \item[] Justification: The paper describes the main information needed to reproduce the experiments, including the quantization setup, bit-budget formula, block size, codebook design, calibration setup, evaluated models and datasets, baseline methods, and implementation details. We also specify the LM Eval Harness versions used for different result tables. In addition, the submission includes a link to the project repository, where we provide the code and supporting materials needed for reproduction.

\item {\bf Open access to data and code}
    \item[] Question: Does the paper provide open access to the data and code, with sufficient instructions to faithfully reproduce the main experimental results, as described in supplemental material?
    \item[] Answer:\answerYes{} 
    \item[] Justification: The submission provides open access to the code and supporting materials, together with instructions for reproducing the main experimental results. The paper also states that the code and supporting materials will be released and describes the main experimental setup, including quantization settings, calibration data, evaluation benchmarks, and implementation details needed to run the experiments.

\item {\bf Experimental setting/details}
    \item[] Question: Does the paper specify all the training and test details (e.g., data splits, hyperparameters, how they were chosen, type of optimizer) necessary to understand the results?
    \item[] Answer: \answerYes{} 
    \item[] Justification: The paper specifies the main experimental details needed to understand the results, including the quantization setup, bit-budget formula, block size, directional and magnitude bit allocations, calibration data, evaluated models and datasets, baseline methods, and implementation details such as the main libraries and evaluation framework versions. These details are provided in Experimental Results and complemented by the appendix tables for additional settings and benchmarks.

\item {\bf Experiment statistical significance}
    \item[] Question: Does the paper report error bars suitably and correctly defined or other appropriate information about the statistical significance of the experiments?
    \item[] Answer: \answerNo{} 
    \item[] Justification: The paper does not report error bars or formal statistical significance estimates for the experimental results. Most evaluations are deterministic once the quantized model is fixed, and the main results are presented as single-run benchmark scores across models and datasets. We instead support the empirical claims through consistent trends across multiple model families, benchmarks, and ablation settings.

\item {\bf Experiments compute resources}
    \item[] Question: For each experiment, does the paper provide sufficient information on the computer resources (type of compute workers, memory, time of execution) needed to reproduce the experiments?
    \item[] Answer: \answerYes{} 
    \item[] Justification:  The paper specifies the main compute resources used in the experiments. Unless stated otherwise, the quantization and evaluation experiments were conducted on a single V100 32GB GPU, while the throughput measurements reported in the appendix were obtained on a single A100 40GB GPU. The appendix also reports tokens-per-second measurements for several Qwen3 models, which provides an additional indication of runtime cost.
    
\item {\bf Code of ethics}
    \item[] Question: Does the research conducted in the paper conform, in every respect, with the NeurIPS Code of Ethics \url{https://neurips.cc/public/EthicsGuidelines}?
    \item[] Answer: \answerYes{} 
    \item[] Justification: The research in this paper is conducted in accordance with the NeurIPS Code of Ethics. The work studies post-training quantization methods for existing language models and does not involve human subjects, crowdsourcing, private data, or any intervention that would raise additional ethical concerns beyond those discussed in the broader impacts section.

\item {\bf Broader impacts}
    \item[] Question: Does the paper discuss both potential positive societal impacts and negative societal impacts of the work performed?
    \item[] Answer: \answerYes{} 
    \item[] Justification: The paper includes a Appendix~\ref{appendix:impact} discussing both a positive impact, namely reduced memory and deployment cost.
    
\item {\bf Safeguards}
    \item[] Question: Does the paper describe safeguards that have been put in place for responsible release of data or models that have a high risk for misuse (e.g., pre-trained language models, image generators, or scraped datasets)?
    \item[] Answer: \answerNA{} 
    \item[] Justification: The paper does not release a new pretrained language model or a new dataset. It studies a quantization method for existing models and may release code for reproducing the experiments, but it does not introduce a standalone high-risk model or data asset that would require additional release safeguards beyond those of the underlying base models.

\item {\bf Licenses for existing assets}
    \item[] Question: Are the creators or original owners of assets (e.g., code, data, models), used in the paper, properly credited and are the license and terms of use explicitly mentioned and properly respected?
    \item[] Answer: \answerYes{} 
    \item[] Justification: The paper explicitly lists and cites the main external assets used in the experiments, and their licenses and terms of use are summarized in Appendix~\ref{appendix:licenses}. For assets whose dataset cards do not clearly specify a license, we state this explicitly and cite the original source.

\item {\bf New assets}
    \item[] Question: Are new assets introduced in the paper well documented and is the documentation provided alongside the assets?
    \item[] Answer: \answerYes{} 
    \item[] Justification: We release the code for the proposed method and provide documentation and usage instructions alongside it. The repository includes the information needed to reproduce the main experiments and describes the scope and intended use of the released implementation.

\item {\bf Crowdsourcing and research with human subjects}
    \item[] Question: For crowdsourcing experiments and research with human subjects, does the paper include the full text of instructions given to participants and screenshots, if applicable, as well as details about compensation (if any)? 
    \item[] Answer: \answerNA{} 
    \item[] Justification: The paper does not involve crowdsourcing, user studies, or any research with human subjects. All experiments are conducted on existing models and benchmark datasets.

\item {\bf Institutional review board (IRB) approvals or equivalent for research with human subjects}
    \item[] Question: Does the paper describe potential risks incurred by study participants, whether such risks were disclosed to the subjects, and whether Institutional Review Board (IRB) approvals (or an equivalent approval/review based on the requirements of your country or institution) were obtained?
    \item[] Answer: \answerNA{} 
    \item[] Justification: The paper does not involve crowdsourcing, user studies, or any research with human subjects, so IRB approval or an equivalent review was not required.

\item {\bf Declaration of LLM usage}
    \item[] Question: Does the paper describe the usage of LLMs if it is an important, original, or non-standard component of the core methods in this research? Note that if the LLM is used only for writing, editing, or formatting purposes and does \emph{not} impact the core methodology, scientific rigor, or originality of the research, declaration is not required.
    \item[] Answer: \answerNA{} 
    \item[] Justification: The core methodology, theoretical results, and experimental pipeline of this work do not rely on LLMs as an important, original, or non-standard component. Any possible use of LLMs for writing or editing does not affect the scientific content of the paper.

\end{enumerate}

\end{document}